# LLM-Generated Negative News Headlines Dataset: Creation and Benchmarking Against Real Journalism

**Olusola Babalola[1]*, Bolanle Ojokoh[2], and Olutayo Boyinbode[3]**

1. Department of Computer Science and Mathematics, Elizade University, Wuraola Ade-Ojo Avenue, P.M.B. 002, Ilara-Mokin, Ondo State, Nigeria; e-mail: olusola.babalola@elizadeuniversity.edu.ng

2. Department of Information Systems, Federal University of Technology, P.M.B. 704 Akure, Ondo State, Nigeria; e-mail: baojokoh@futa.edu.ng

3. Department of Information Technology, Federal University of Technology, P.M.B. 704 Akure, Ondo State, Nigeria; e-mail: okboyinbode@futa.edu.ng

**Abstract**

This research examines the potential of datasets generated by Large Language Models (LLMs) to support Natural Language Processing (NLP) tasks, aiming to overcome challenges related to data acquisition and privacy concerns associated with real-world data. Focusing on negative valence text, a critical component of sentiment analysis, we explore the use of LLM-generated synthetic news headlines as an alternative to real-world data. A specialized corpus of negative news headlines was created using tailored prompts to capture diverse negative sentiments across various societal domains. The synthetic headlines were validated by expert review and further analyzed in embedding space to assess their alignment with real-world negative news in terms of content, tone, length, and style. Key metrics such as correlation with real headlines, perplexity, coherence, and realism were evaluated. The synthetic dataset was benchmarked against two sets of real news headlines using evaluations including the Comparative Perplexity Test, Comparative Readability Test, Comparative POS Profiling, BERTScore, and Comparative Semantic Similarity. Results show the generated headlines match real headlines with the only marked divergence being in the proper noun score of the POS profile test.

## 1.0. Introduction

Our experiments aim to provide data and insights into Large Language Model (LLM)-generated negative content. We first generate synthetic news headlines using a chosen LLM and across specific and general categories; we assess how well each headline bears negative sentiment and analyze how negativity varies across different news categories; finally, we examine patterns and characteristics in the LLM-generated negative headlines. The experiments are expected to produce several useful results, including comparing how different sentiment analysis tools perform on negative news headlines, understanding the strengths and weaknesses of various sentiment analysis methods across different news categories, seeing how well current sentiment analysis tools detect negative sentiment in AI-generated content, finding any biases or patterns in how LLMs create negative news, and showing how these findings might affect the use of these tools in real-world applications, especially for news monitoring and analysis.

News headlines play a crucial role in shaping public opinion and perception, and a lot of utility is being derived from the ability to use computers to accurately detect negative sentiment in news. Having a broader understanding of these sentiments as well is of great importance to various stakeholders, including media analysts, policymakers, and researchers. Using LLM-generated content as dataset, we position this research at the intersection of two rapidly evolving fields: sentiment analysis and large language models. This approach not only provides a controlled environment for testing sentiment analysis tools but also offers insights into the nature of AI-generated news content, a topic of increasing relevance as AI systems become more integrated into content creation pipelines. By focusing on negative sentiment detection in news headlines, this research addresses an important area of sentiment analysis with significant real-world implications.

The findings from this research have the potential to inform the development and refinement of sentiment analysis tools, particularly for applications in news analysis. The insights gained about LLM-generated negative content could contribute to ongoing discussions about AI ethics, media literacy, and the potential impacts of AI-generated news on public discourse. We proceed to discuss negativity in news, synthetic datasets, and LLMs.

## 2.0 Related work

Synthetic data, which is simply data generated by algorithms that is similar in every way to real world data but without direct derivation from actual, or naturally occurring objects or entities. They are therefore purely the creation of models which have some knowledge of the world the data is being generated for. The primary use of synthetic data is for the training of machine learning systems. The literature definition of synthetic data includes the usage of machine generated data to train a model [1] However, we are witnessing other usage of model-generated datasets especially those from generative A.I. (genAI). An apparent advantage of generating data using LLM is availability, a significant reduction in time and effort required for data collection and labelling, and the low or (almost) zero risk of privacy breaches or invasion [2]. Synthetic data also offers a stable and known body of truth, [3] designed an algorithm to generate synthetic text data containing "an unambiguously defined ground truth structure" for the purpose of testing the performance of probabilistic topic modeling algorithms [3]. They thereby achieve an evaluation method that is formal and objective in every way for the topic model evaluation problem by using synthetic data.

Sentiment analysis, a crucial aspect of natural language processing, faces significant challenges due to data imbalance and scarcity issues [4]. Traditional data collection methods, such as annotators, are not only time-consuming and expensive but also prone to bias [5]. To address these limitations, researchers are drawn to synthetic text generation techniques. Recent studies have demonstrated the effectiveness of Generative Adversarial Networks (GANs) in generating high-quality synthetic text for sentiment analysis. For instance, CatGAN and SentiGAN have been employed to balance highly imbalanced datasets, resulting in significant performance improvements in sentiment classification tasks [4], highlighting the potential of GAN-based models in handling data imbalance issues. Novel approaches combining transformers and evolutionary algorithms have also shown promise in generating diverse and accurate synthetic data for sentiment analysis [5]. The SentiGEN system, leveraging T5 and XLNet, demonstrated superior performance compared to traditional models trained on real data [5] underscoring the importance of synthetic generation in solving data-related challenges.

Quality issues arise when dealing with the generation of text in situations where there is limited data or low-resource languages. Recent studies have shown that sophisticated models like GPT and BERT, reliant on NLP transformers, can generate creative text but struggle with proper sentence structure when data is scarce [6]. To address this challenge, the authors explored novel approaches focusing on post-processing techniques and linguistic feature-based methods. They used GPT-2 augmented with Part-of-Speech (POS) tagging patterns to enhance text generation quality in their news headline generation tasks.

But how does algorithm-generated text data fare in language processing scenarios? We believe it is important to understand the treatment of computational processes largely designed to analyze human-produced content when they encounter model-produced content. A research study found algorithms were more biased towards detecting fake news from a LLM than those produced by humans [7]. The case may be different in the information retrieval (IR) research field. Ref [8] conducted a study to investigate information retrieval (IR) systems in the new setting involving the retrieval of model-created content, and human-created content stored in the same index [8]. Their findings showed neural retrievers (in short, algorithms) favored content produced by LLMs (also algorithms). Another question in the use of model-generated data in text form is what differences exist between the text generated by LLMs and those produced by humans? Research shows discernible differences in morphological, syntactic, psychometric, and sociolinguistic aspects [9]. Outputs of models trained using LLM-generated data have also shown diverse performances [10].

An advantage obtained in using LLMs for data generation is their ability to follow instructions (prompts) to produce diverse and faithful data through heuristic prompting [11]. We are interested in LLaMA 3, an LLM from Meta AI, for use in generating news headline dataset. The Meta AI LLaMA 3 model is an LLM artificial intelligence system built on a transformer-based neural network architecture. The model has been trained on a large-scale corpus of text sourced from various domains, including web pages, books, articles, research papers, and online forums; it has over 70 billion parameters with capabilities in contextual understanding, natural language processing (NLP), and natural language generation (NLG). LLaMA 3 is used for answering questions, generating text, translating, summarizing, and engaging in conversational dialogue. The model's limitations are like those of other LLMs in its category including, limited domain-specific knowledge, the absence of personal experiences or emotions, dependence on the quality and biases of training data, and potential challenges in understanding nuances or context. The LLaMA 3 base model was released in 2023.

However, OPT-175B, another open Meta AI LLM with 175-billion parameter model, may be an ideal choice for use cases involving generating large datasets due to its unique training approach that prioritized transparency and reproducibility in AI research [12]. The model was trained on a diverse corpus of web content, books, and social media posts, making it particularly suitable for generating news-like content. LLaMA 3 models, though typically smaller in size than OPT-175B, focus on efficiency and accessibility for researchers who may not have access to huge computational resources. While OPT-175B prioritizes open, large-scale experimentation, LLaMA 3 models are designed to be lightweight yet powerful, catering to resource-constrained environments while still delivering robust performance.

How do these models compare to the closed GPT-4 models from Open AI? OPT-175B model, differs from GPT-4 in terms of openness and accessibility. While GPT-4 operates within a more controlled environment, OPT-175B is explicitly designed for open research. Meta AI provides detailed logs, training processes, and model weights for this model, allowing researchers to freely explore and experiment with its capabilities in AI development and study. Performance wise, LLaMA 3 compares favorably in terms of quality to leading language models such as GPT-4 on a wide variety of tasks [13].

Using a large language model (LLM) trained on a substantial collection of public content to generate a dataset of negative news across general and specified categories provides a fine-grained reflection of real-world issues. Most general LLMs, already informed by patterns in journalistic output, can generate news headlines that mirror the style and structure typical of everyday media while understanding sentiment distribution thereby making them capable of producing realistic fake news [14]. This motivates us to use them as a method that ensures the production of a diverse and representative dataset offering a detailed approximation of the complexities found in actual negative news stories and having the sentiment controlled to suit specific analytical goals. LLMs are highly effective in generating vibrant, contextually sound and applicable text content in different domains. Their advanced language understanding capability further enhances their capability in managing tasks requiring contextual understanding and response towards achieving required text outputs [15]. Once running, a LLM responds almost instantly to minimal prompting, supplying contextually generated data satisfying the input prompt. The prompt may be modified to give feedback on the data if necessary. This is advantageous to users who may need a particular type of text data.

Key attributes of LLMs include few-shot learning, contextual awareness across diverse subject matter, and enhanced control over content bias. Few-shot learning in large language models (LLMs) allows models to generate high-quality content with minimal examples. This has been explored in several studies, notably by [16] which highlights how few-shot learning reduces the need for retraining while maintaining strong performance. Also, transformer-based architectures like LLaMA 3 are designed to capture long-range dependencies and contextual relationships. This attribute is a core principle outlined in [17]. Another attribute is bias control; handling bias in AI-generated text is of interest to the research community, particularly in synthetic data generation which outputs often end up being used to train AI models. Techniques like pre-processing and heuristic strategies to mitigate bias and ensure balanced outputs have been discussed in literature for instance in [18].

Employing LLM-generated content fulfills several objectives: It supplies a steady stream of headlines across diverse categories, mitigating potential biases from disparate news sources or writing styles; it enables us to investigate how LLMs interpret sentiments and produce negative news thereby providing valuable insights into potential biases or patterns in AI-generated content; it establishes an innovative testbed for sentiment analysis tools, posing challenges with content that closely resembles human-written headlines yet originates from an AI. But it is not without issues. The proliferation of large language models (LLMs) has led to an increased reliance on artificially generated data, raising concerns about quality, diversity, and potential biases [19]. In that regard, recent studies have highlighted the limitations of LLM-generated data; issues particularly in capturing human traits, subtleties of language, and minority viewpoints. Despite demonstrating human-like performance in various tasks, LLMs have been found to struggle to replicate the complexity of human language structures and styles. Ref [19] stress-tested LLM-generated data across various benchmarks, revealing significant disparities with human-generated content, especially in intricate and subjective tasks.

## 3.0. Materials and methods

The methodology to achieve our research objectives centers on three crucial elements; the generation of a negative news dataset, examining the generated headlines for realism using distance measures and also a POS-based evaluation used in [6], and a multi-faceted sentiment analysis of the generated dataset.

### 3.1. Dataset generation

The synthetic data generation pipeline leverages the pathway obtained from [20] which provides an extensive workflow of how large language models (LLMs) is used in generating, curating, and evaluating synthetic data [20]. The framework also addresses challenges like ensuring diversity in synthetic data and the need for domain-specific expertise to guide LLMs in generating accurate and contextually relevant data. It emphasizes leveraging large language models (LLMs) to create high-quality (meaning it which passes faithfulness, and diversity yardsticks) synthetic data with specific characteristics. In summary, the framework involves three key activities: data generation, data curation, and data evaluation.

The data generation process entails harnessing Large Language Models (LLMs) to produce synthetic data. Key activities at this state of the workflow encompass prompt engineering which involves formulating effective prompts that outline tasks, generation conditions, and in-context demonstrations to guide LLMs in generating coherent and pertinent outputs; multi-step generation which entails implementing a structured approach where data is created in successive stages enabling iterative refinement and enhancement of the generated samples; sample-wise and dataset-wise decomposition whereby complex generation tasks are divided into manageable components to improve the quality and relevance of the generated data.

Data curation concentrates on refining the generated data to guarantee high quality and relevance. Key activities involve sample filtering by employing heuristic metrics to identify and eliminate low-quality samples based on confidence scores and relevance to the desired labels; label enhancement for enhancing the accuracy of labels through manual re-annotation and auxiliary model improvement to minimize noise and errors in the dataset; active selection by engaging human experts to review and select high-quality samples, ensuring that the curated dataset fulfills the necessary criteria for downstream tasks.

The data evaluation phase examines the quality and efficacy of the generated and curated data. Key activities at this stage are direct evaluation involving comparison of generated samples to ground truth data when accessible, to assess data faithfulness and coherence; human evaluation which leverages human experts to evaluate the quality of the generated data based on predefined criteria, providing insights into the overall generation quality; benchmark evaluation through conducting systematic assessments using established benchmarks to evaluate the synthetic data's performance in various tasks, ensuring its suitability in real-world scenarios.

Following the framework's curation stage, we implemented a two-step filtering process involving filtering using predefined rules to remove duplicates but did not filter out nonsensical headlines and only used human validation to evaluate if the generated headlines met quality standards and truly reflected negative sentiment.

### *3.1.1 LLM for news dataset generation*

Given its previously examined capabilities, LLaMA 3 is well-suited for this research, where it is used to produce negative news headlines that mirror the diversity and complexity of real-world journalism while maintaining control over the sentiment. LLaMA 3's ability to handle long-range dependencies and generate coherent text aligns with advancements discussed in transformer-based language models e.g. BERT [20].

### *3.1.2 Process of news generation and validation*

The news headline generation process follows a structured approach based on the LLMs-driven framework. We began with a task specification, defining a clear context and instruction set for the LLM. In this study, the goal is to generate news headlines dataset with negative sentiment. This goal's data generation phase was broken down into multiple stages. Stage one was for the LLM to generate the headlines according to its own determined categories. Stage 2 was for the researcher to generate the headlines across 35 common news categories (e.g., politics, business, lifestyle). At any point during any of these stages, the researcher had liberty to modify the prompt if the output by the LLM was not satisfactory, and based on this premise there was ground in Stage 1 in this case to modify along generation runs to improve the diversity of generated headlines. A checklist of synthetic data generation based on the reference work [20] is presented in Figure 1.

**Data Generation**

- **Prompt Engineering**
  - Design task specifications (e.g., task purpose, data explanation).
  - Define generation conditions (e.g., role-play, format clarification).
  - Create in-context demonstrations (e.g., examples of desired outputs).
- **Multi-Step Generation**
  - Identify sub-tasks for multi-step generation.
  - Schedule generation procedures for each sub-task.
  - Generate intermediate outputs using model prompts.
- **Sample-Wise Decomposition**
  - Break down complex samples into smaller chunks.
  - Ensure coherence among generated components.
  - Implement Chain-of-Thought (CoT) prompting for reasoning.
- **Dataset-Wise Decomposition**
  - Generate data with specified properties for diversity.
  - Create a series of samples to form a comprehensive dataset.

**Data Curation**

- **Sample Filtering**
  - Apply heuristic metrics to evaluate sample quality.
  - Discard low-quality or irrelevant samples based on confidence scores.
- **Label Enhancement**
  - Conduct manual re-annotation of samples for accuracy.
  - Utilize auxiliary models to improve label quality.
- **Demonstration Selection**
  - Acquire relevant demonstrations for task-specific guidance.
  - Select high-quality demonstrations for inclusion in the dataset.
- **Human Intervention**
  - Engage human experts for annotation and verification.
  - Ensure readability and interpretability of LLM outputs.

**Data Evaluation**

- **Direct Evaluation**
  - Compare generated samples against ground truth data.
  - Assess data faithfulness and coherence.
- **Human Evaluation**
  - Involve human experts to evaluate sample quality.
  - Establish criteria for quality assessment.
- **Benchmark Evaluation**
  - Conduct systematic assessments using established benchmarks.
  - Evaluate the performance of synthetic data in various tasks.
- **Feedback Loop**

- Gather feedback from evaluations to inform future generations.
- Adjust prompt engineering and generation strategies based on insights.

Figure 1: Checklist for LLMs-Driven Synthetic Data Generation Workflow

### *3.1.3 Prompt Engineering: prompting data generation*

The LLM is provided with carefully crafted prompts designed to elicit negative news headlines. The key properties of the prompts follow the structure in Figure 2. This ensures that the resulting dataset captures the negative sentiment while remaining structurally relevant to most new categories which we have found to average about 8 words in length. A specific prompt is provided to the LLM to generate negative news headlines. The master prompt which is used to kickstart the major objective is crafted to have intent or objective of what we want to use the data being generated for, properties and qualities of the data to be generated, we specify the form, the structure, the sense, etc., and the number of data points to be created. An analysis of the structure of an example prompt is provided in Figure 2.

**Data Generation:**

- **Type:** Text
- **Structure:** Tabular or free-flow
- **Rows:** The number of headlines to generate, between 30-50
- **Features:** Non-stopwords, negative valence biased statements
- **Use Case:** a model to detect general negative news headlines with over 99% accuracy.
- **Average length per Headline:** 8 words
- **Bias:** Negative valence, general negativity (not specific to any group or segment of society)

Figure 2: Prompt Structure Sample I

### *3.1.4 Controlled sentiment distribution*

The LLM model generated headlines in about 58 categories, with 5 – 15 negative news headlines per category. The categories range from politics and technology to social issues, thus ensuring a wide range of sentiment-focused content. To broaden the diversity of the dataset, the iterative synthetic data generation process started with the master prompt, and an iterative prompting as the LLM returns results. With each data generation batch, a prompt is given to instruct the LLM continue generation while maintaining the previous configuration which included the valence objective.

### *3.1.5 Controlled news category distribution*

The LLM is allowed to show its category bias by asking it to generate headlines without specifying or restraining categories. Diversification of the dataset is, however, pursued by requesting generation of outputs to cover previously unexplored segments as illustrated in the set of prompts in Figure 3. The prompts instructing the model to generate new headlines in a diverse manner in each subsequent prompt. The LLM is iteratively (re)prompted to broaden the scope of its outputs. A comprehensive and diverse dataset is effectively generated, incorporating various constraints and directions to guide the process. The LLM was allowed to freely generate headlines in the first stage, it had the liberty of determining what categories of headlines to generate. A total of 28 categories were covered by the LLM.

- Generate 50 fresh headlines (according to the master prompt)
- Generate 50 headlines in *diverse segments of life and society*, ensuring a broader scope of negativity.
- Generate 50 headlines, maintaining the negative valence bias while *exploring other areas of interest* you have not covered, expanding the scope of the generated data.
- Generate 50 headlines *with an international bias*, introducing a new dimension of negativity and highlighting global issues.
- Generate fresh headlines in segments that had not been previously covered, further diversifying the data generated and increasing the overall complexity of the dataset.
- Continue this iterative process, repeatedly instructing the model to generate new headlines, with each subsequent prompt introducing a new set of constraints or directions, effectively guiding the generation of a comprehensive and diverse dataset.

Figure 3. Multi-step generation / sample-wise decomposition / chain of thought

To generate a comprehensive dataset of negative news headlines, a chain of thought approach was employed, as was recommended in the framework. This involved a series of iterative prompts, each designed to guide the model towards a specific aspect of negativity and diversity. Firstly, the model was prompted to generate 50 headlines, this is repeated multiple times with the LLM focusing on general negative events. After looping through multiple times without modifying the prompt, this provided a baseline for the dataset. Follow up prompts were well utilized. Following the master prompt that kickstarted the generation of 50 news headlines, subsequent generation runs involved a series of follow-up prompts. These prompts were structured in a pattern that iteratively guided the synthetic data generation process.  Specifically, each subsequent run instructed the model to refine and expand the dataset. Examples of the prompts can be found in Figure 3. The model was constantly prompted to generate more headlines with necessary guidelines for example requesting it generates negative news headline but this time with a focus on diverse segments of life and society. This ensured that the dataset captured a wider range of negative events, from economic downturns to social unrest, encompassing various aspects of human experience. In line with the objective, the model was prompted to generate more headlines, with a specific focus on the LLM's chosen areas of interest further diversifying the dataset and ensuring the dataset is relevant for the intended application. The process continued with prompts that introduced international bias, highlighting global issues and events, and then prompts that focused on areas that had not been previously covered, further expanding the scope and complexity of the dataset. At the end of stage 1, about 150 news headlines were generated.

#### *3.1.6 Constrained news category distribution*

The next stage involved the researcher prompting specifically a set of predefined categories. This allowed us to explore specific themes or topics relevant to our research goals. There were 35 pre-selected categories and the LLM was prompted to generate negative news headlines for all. It is important to note that the initial set of negative news headlines generated in Stage 1 before prompt modifications show a skew towards some categories (See Figure 3), also the categories obtained from the headlines generated in the second stage therefore had overlaps with our predefined categories.

By following this chain of thought approach, we were able to create a comprehensive and diverse dataset of negative news headlines that captured a wide range of negativity and addressed different aspects of human experience and global events. This approach ensured that the dataset was suitable for training a model to detect general negative news headlines with high accuracy. The research objective of creating a comprehensive and diverse dataset of news headlines of negative valence bias, covering various segments of life and society, specific areas of interest, and international issues, was systematically achieved ensuring that the generated data encompassed a broad spectrum of negativity and covered diverse aspects of human experience and global events.

#### *3.1.7 Output curation*

As the LLM returned the generated news headlines, they were inspected by one of the authors who was tasked to examine format, closeness to reality, structure, valence, bias, and coverage. The author was to modify the prompt if needed to align with the goals of the initial task.

After the headlines are generated, they are curated by checking for content that does not align with the desired sentiment or contain inaccuracies. Manual inspection and sentiment verification ensure that the headlines adhere to journalistic standards and reflect real-world negativity in news reporting. In our case the major curation task was the removal of similar headlines. The curation was therefore partly done in real time by modifying prompts as well as after the generation was done by examining the overall dataset. The curation and refinement process ensures headlines meet a category spread and that repetitive headlines are curated out of the dataset, maintaining both diversity and accuracy.

## 3.2 Validation through Sentiment Analysis

Once the headlines are generated, they are validated using established sentiment analysis tools. This step evaluates how well the generated headlines reflect the intended negative sentiment. Multiple sentiment analysis methods, popular in real-world applications, are employed to ensure robustness and provide comparative insights across different algorithms. The curated headlines are then validated using popular sentiment analysis assessing whether the sentiment analysis tools can detect the intended negative sentiment. This step helps verify that the data generated is both valid for sentiment analysis research and reflective of actual news text.

We present a custom checklist for synthetic data generation workflow using LLMs for the generation of news headlines in Figure 4.

**Data Generation**

- **Prompt Engineering**
  - **Design task specifications**: Define the objective to generate negative headlines for a specific topic (e.g., movie reviews, product feedback).
  - **Define generation conditions**: Specify the tone (e.g., critical, sarcastic) and length of the headlines (e.g., 5-10 words).
  - **Create in-context demonstrations**: Provide examples of negative headlines to guide the model (e.g., "Train hits transport bus, scores injured").
- **Multi-Step Generation**
  - **Identify sub-tasks**: Break down the task into generating different types of negative headlines (e.g., reviews, news articles).
  - **Schedule generation procedures**: Plan to generate a batch of headlines, review them, and refine the prompts if necessary.
  - **Generate intermediate outputs**: Use the model to produce initial sets of negative headlines based on the prompts.
- **Sample-Wise Decomposition**
  - **Break down complex samples**: If generating longer headlines, divide them into key components (e.g., subject, critique).
  - **Ensure coherence**: Check that each generated headline maintains a consistent negative sentiment.
  - **Implement Chain-of-Thought (CoT) prompting**: Guide the model towards attaining goals in sequential prompts.
- **Dataset-Wise Decomposition**
  - **Generate data with specified properties**: Create a diverse set of negative headlines across various topics (e.g., politics, environment).
  - **Create a series of samples**: Aim for a comprehensive dataset that captures different negative sentiments and styles.

**Data Curation**

- **Sample Filtering**

- **Apply heuristic metrics**: Evaluate the generated headlines for negativity and relevance.
    - **Discard low-quality samples**: Remove headlines that are vague, overly generic, or do not convey a negative sentiment.
- **Label Enhancement**
    - **Conduct manual re-annotation**: Review the remaining headlines to ensure they accurately reflect negative sentiment.
    - **Utilize auxiliary models**: Use sentiment analysis tools to verify the negativity of the headlines.
- **Demonstration Selection**
    - **Acquire relevant demonstrations**: Gather high-quality examples of negative headlines from existing literature or media.
    - **Select high-quality demonstrations**: Choose the best examples to serve as benchmarks for future generations.
- **Human Intervention**
    - **Engage human experts**: Have editors or content creators review the headlines for clarity and impact.
    - **Ensure readability**: Confirm that the headlines are easily understandable and effectively convey negativity.

**Data Evaluation**

- **Direct Evaluation**
    - **Compare generated samples**: Assess the generated headlines against a set of known negative headlines for consistency.
    - **Assess data faithfulness**: Ensure that the headlines accurately reflect the intended negative sentiment.
- **Human Evaluation**
    - **Involve human experts**: Have reviewers evaluate the quality of the headlines based on criteria such as creativity and negativity.
    - **Establish criteria**: Define what constitutes a successful negative headline (e.g., clarity, emotional impact).
- **Benchmark Evaluation**
    - **Conduct systematic assessments**: Compare the generated headlines to a benchmark dataset of negative headlines.
    - **Evaluate performance**: Analyze how well the headlines generated perform in terms of engagement and sentiment.
- **Feedback Loop**
    - **Gather feedback**: Collect insights from evaluations to identify strengths and weaknesses in the headlines generated.
    - **Adjust prompt engineering**: Refine prompts and generation strategies based on feedback to improve future headline generation.

Figure 4: Synthetic negative news headlines dataset generation checklist based on [20]

## 3.3 Dataset Size and Characteristics

The final dataset comprising 510 negative news headlines were generated through Stage 1 and Stage 2. Further examination showed certain headlines were repeated, after removal of repeated headlines, 467 unique headlines remained. The distribution of the headlines can be found in Figure 5 and Appendix A1. These headlines span categories such as politics, economics, environmental issues, rights, and technology, ensuring a wide scope of real-world applicability. Each headline is designed to reflect a strong negative sentiment, tailored to capture the often-ambiguous nature of negativity in news. The average length of each headline is under 10 words, adhering to the typical format of impactful, concise news reporting. For each of the 35 news categories, approximately 5-10 negative news headlines were generated. The categories span a wide array of topics. Figure 5 shows the categories generated by the LLM and Figure 6 the categories we requested.

Each headline is a concise, impactful statement designed to mimic the style of mainstream news outlets, with attention to maintaining a realistic distribution of negativity across the dataset. Characteristics of the dataset include standard length, negative tone across all categories, diversity, balanced distribution and sentiment control. The headlines were all under 10 words in length, all headlines reflect a negative sentiment, focusing on topics such as economic downturns, political instability, environmental disasters and similarly, the domain issues in each category. The dataset exhibited diversity as headlines cover a wide variety of negative news events, ensuring that the dataset captures the complexity and reality typical of real-world journalism. The dataset contained headlines in 58 categories with each category having at least five (5) negative headlines biased to the category, ensuring that the dataset is evenly distributed across topics. The fit was confirmed by a human evaluator.

The actual amount of news in each category is relative to its popularity in the real world. The prompts generating the headline maintained sentiment control whereby headlines are created to reflect strong negative sentiment, with sentiment validation ensuring that no neutral or positive headlines are included. And finally, the dataset has topical diversity; all headlines share a negative tone, and the subject matter spans a diverse range of societal issues, providing a broad basis for testing sentiment analysis tools if so desired.

The target categories are shown in Figures 5 and 6; a randomly chosen set of categories are visualized as wordclouds in Figure 7 and more in Appendix A3. A breakdown of the number of headlines generated per category is presented in the table in Appendix A1 and visualized in Figure 8.

| | | | | | |
|---|---|---|---|---|---|
| *News* | *Travel* | *Regional* | *Lifestyle* | *Feature* | *Cooking* |
| *Local* | *Technology* | *Press releases* | *Home* | *Entertainment* | *Comics* |
| *Journalism* | *Sports* | *Politics* | *Health* | *Education* | *Celebrity* |
| *Government* | *Society* | *Organizations* | *Gardening* | *Editorials* | *Business* |
| *World* | *Social* | *Obituaries* | *Frontpage* | *DIY* | *Auto* |
| *Weather* | *Science* | *National* | *Headline* | *Crime* | |

Figure 5: Custom requested news categories

| | | | | | |
|---|---|---|---|---|---|
| *News* | *Travel* | *Regional* | *Lifestyle* | *Feature* | *Cooking* |
| *Local* | *Technology* | *Press releases* | *Home* | *Entertainment* | *Comics* |
| *Journalism* | *Sports* | *Politics* | *Health* | *Education* | *Celebrity* |
| *Government* | *Society* | *Organizations* | *Gardening* | *Editorials* | *Business* |
| *World* | *Social* | *Obituaries* | *Frontpage* | *DIY* | *Auto* |
| *Weather* | *Science* | *National* | *Headline* | *Crime* | |

Figure 6: News categories generated by LLM

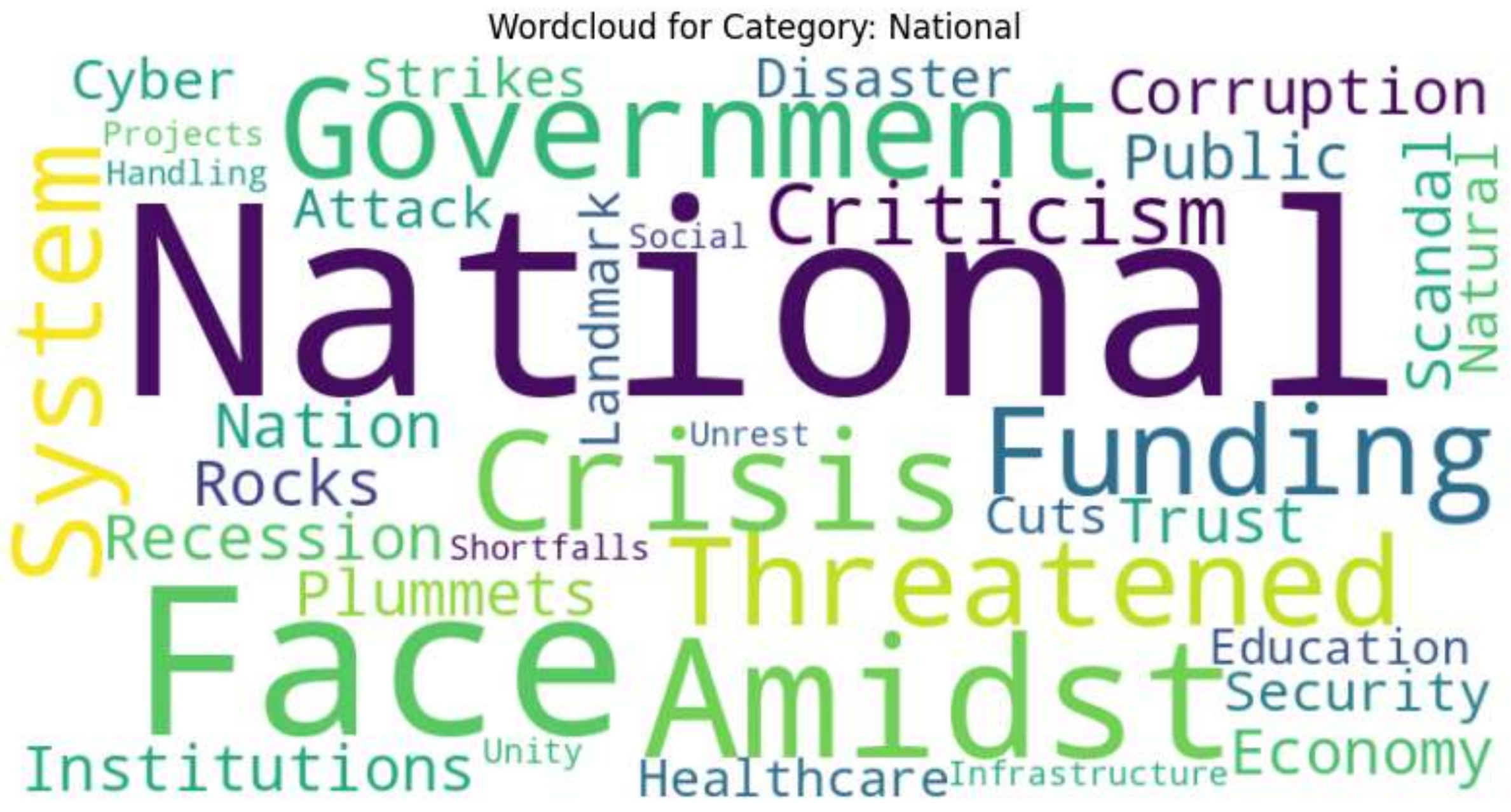

Figure 7a. Wordcloud for National category

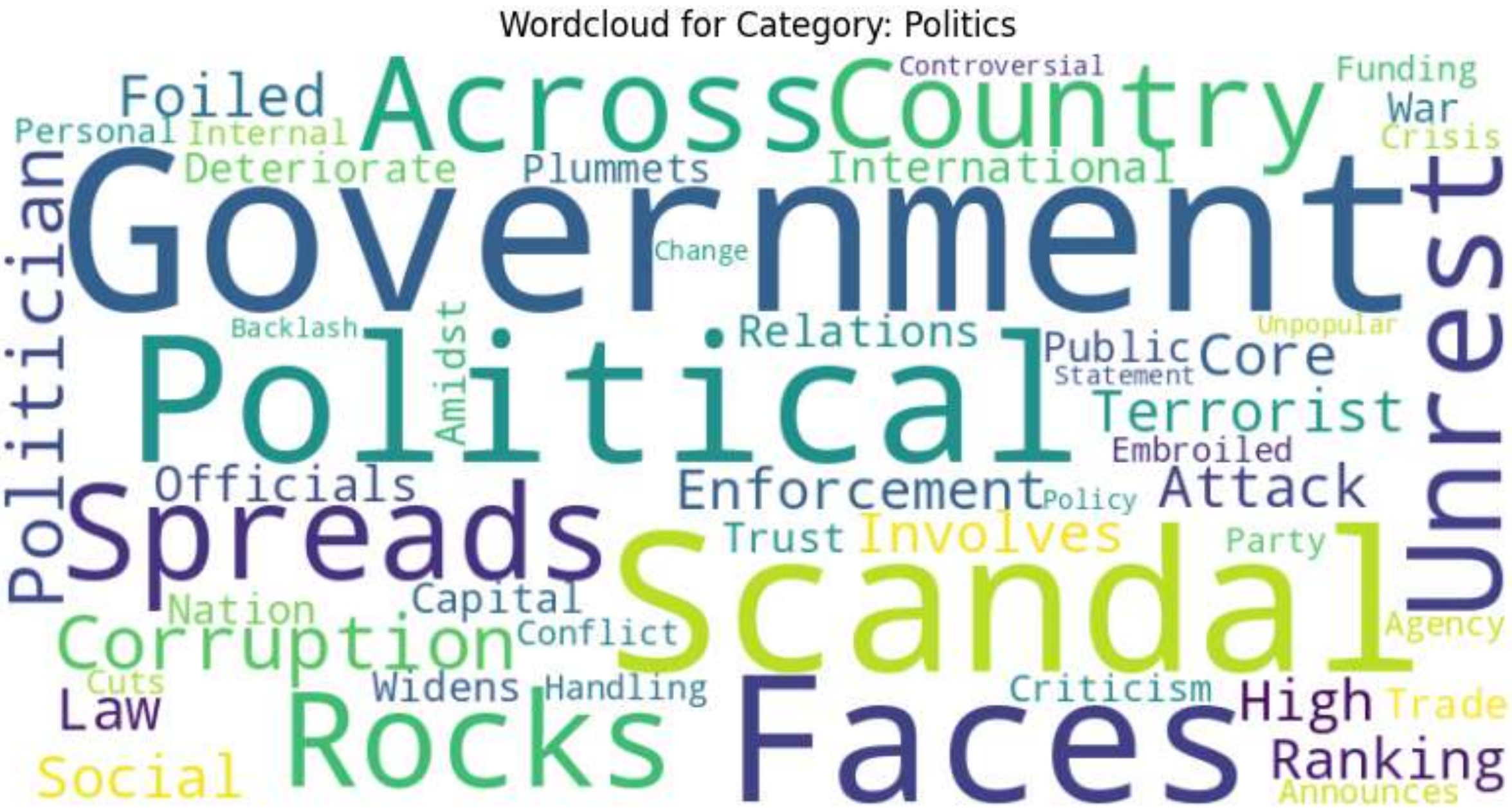

Figure 7b. Wordcloud for Politics category

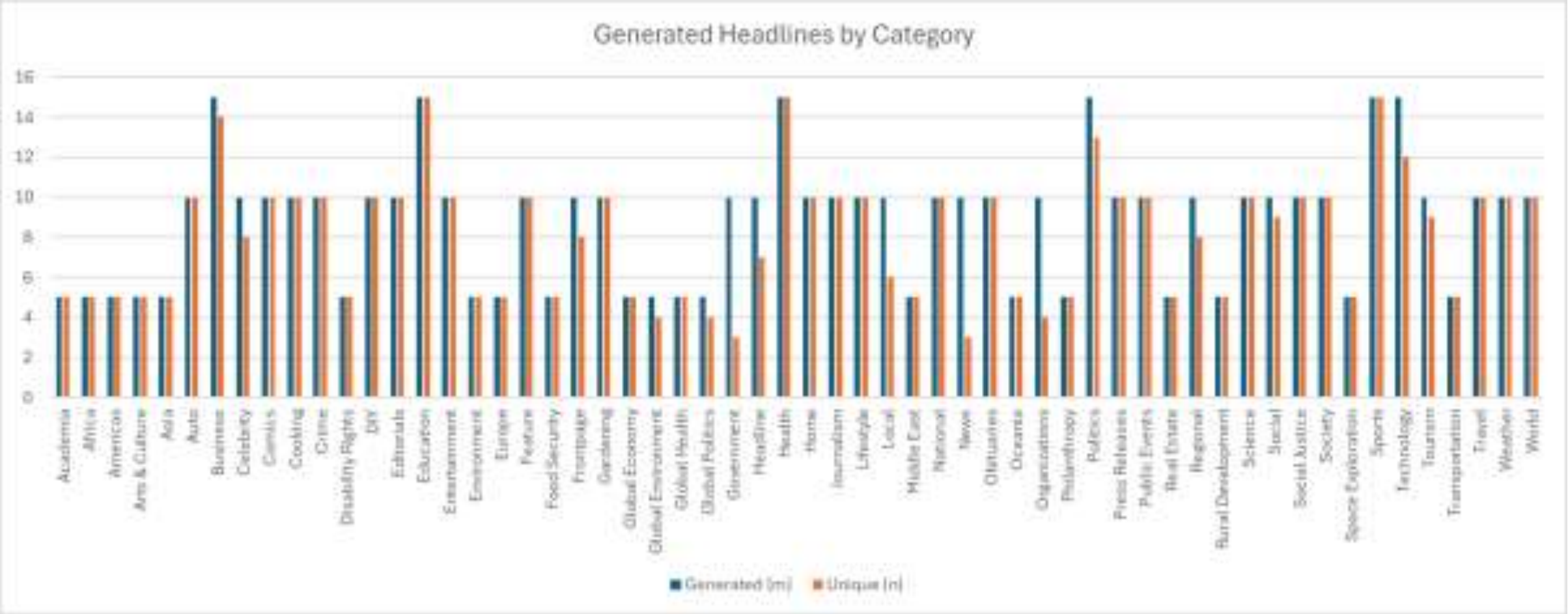

Figure 8. Generated headlines and unique headlines

## 3.4 Realism evaluation of the synthetic news corpus

We undertake some assessments to determine how authentic and what the quality and realism of the generated headlines are by an evaluation methodology that moves beyond traditional metrics used in text summarization or generation, such as Recall-Oriented Understudy for Gisting Evaluation (ROUGE),

Bilingual Evaluation Understudy (BLEU), and Metric for Evaluation of Translation with Explicit Ordering (METEOR). Although these metrics have been used in literature [21], but those metrics mostly rely on direct n-gram overlap and they fall short in capturing the dynamism of human language. This is especially so in a domain-specific context such as news writing, and are therefore less effective when evaluating unaligned or diverse content [22], [23]. We opted for a more comprehensive approach that focuses on corpus-level evaluation using perplexity scores, semantic similarity measures, and readability metrics. This allows us to assess the headlines' fluency, coherence, and alignment with human writing. To assess the performance of Large Language Models (LLMs), a combination of metrics is frequently employed to furnish a comprehensive and multidimensional perspective on their efficacy across a diverse range of tasks [24] - BLEU and METEOR metrics enables the evaluation of machine translation models; ROUGE is typically leveraged to assess the quality of summarization tasks, and BERTScore is exploited to evaluate tasks necessitating semantic analysis, thereby providing a fine understanding of the models' capabilities in capturing semantic subtleties and contextual intricacies

To evaluate the quality of the generated sentences, we collected a set of 10815 from a dataset of human-written news sentences from various reputable newspapers [25], the headlines are from over 100 categories. This collection was manually examined and cleaned to ensure that it represents a broad range of topics and styles typically found in journalistic writing. Importantly, the human-written headlines were not directly aligned with the LLM-generated sentences in terms of content. While both sets of sentences are news-related, there was no attempt to match each generated sentence with a corresponding human-written sentence, as our goal was to conduct a general corpus-level evaluation of the generated text. A distribution of the headlines in the confirmation dataset is presented in Appendix A1.

### 3.4.1 Generated headlines quality checking via corpus-level examination

A comprehensive evaluation was sought by focusing on fluency, semantic content, sentence stylometry, and readability, factors for determining the realism and quality of news headlines. Figure 9 illustrates our evaluation steps; the steps are not necessarily ordered nor are they a pipeline. Overall, this approach offers several advantages over traditional n-gram based similarity metrics. Firstly, it accounts for semantic equivalence even when different words are used to express similar concepts. Secondly, it is more robust to variations in sentence structure and word order as it is focused on the overall meaning conveyed by the headline. Lastly, by using a pre-trained model fine-tuned on a diverse corpus, we leverage transfer learning to capture fine language understanding that might be missed by simpler techniques. This comprehensive evaluation approach offers insights into the stylistic and structural characteristics of the headlines, moving beyond simple word-level comparisons to consider broader patterns in language use. Such an approach provides an understanding of the differences in headline composition between the two datasets, potentially reflecting variations in journalistic style, content focus, or target audience. By integrating these evaluation methods, we are able to overcome the limitations in traditional metrics that rely on direct n-gram overlap, which does not effectively measure the quality of content when the generated text is not a direct paraphrase or translation of a reference text [22], [23].

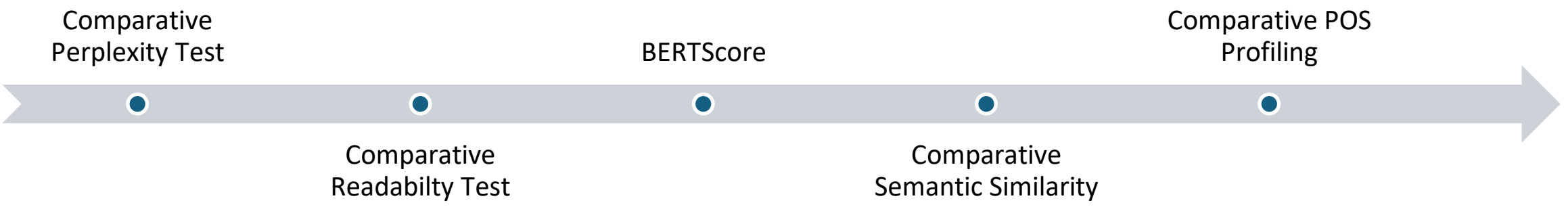

Figure 9. Evaluation process

### 3.4.2 Perplexity

We undertook a comparative perplexity test, calculating perplexity scores using a trigram language model with Kneser-Ney smoothing [26]. Perplexity measures how well a probability model predicts a sample. In our context, perplexity gauges the fluency and grammatical correctness of the headlines. Lower perplexity scores indicate that the headlines are more predictable and align closely with standard language usage found in news reporting. The average perplexity score of generated headlines is compared to that of human-written headlines; if they are comparable, it suggests the generated headlines maintain a high level of fluency and adhere to grammatical norms typical of professional journalism.

Perplexity is computed as shown in Equation 1.

$$Perplexity = \exp\left(-\frac{\sum \log P\,(w_i|w_{i-2}, w_{i-1})}{N}\right) \tag{1}$$

where $P(w_i|w_{i-2},w_{i-1})$ is the conditional probability of the current word $w_i$ given the previous two words $w_{i-2}$ and $w_{i-1}$, and $N$ is the number of words in the headline.

The conditional probability $P(w_i|w_{i-2},w_{i-1})$ is calculated using the modified Kneser-Ney smoothing method, as shown in Equation 2:

$$P(w_i|w_{i-2}, w_{i-1}) = \frac{\max(\text{count}(w_{i-2}, w_{i-1}, w_i) - d, 0)}{\text{count}(w_{i-2}, w_{i-1})} + \lambda_{w_{i-2},w_{i-1}} \cdot P_{\text{continuation}}(w_i|w_{i-1}) \tag{2}$$

where $count(w_i - 2, w_i - 1, w_i)$ is the trigram count, $\mathrm{count}(\mathrm{w_{i-2}}, \mathrm{w_{i-1}})$ is the bigram count, *d* is the discount parameter (set to 0.75), $\lambda_{\mathrm{w_{i-2}},\mathrm{w_{i-1}}}$ is the interpolation weight, and $\mathrm{P_{continuation}}(\mathrm{w_i} | \mathrm{w_{i-1}})$ is the continuation probability.

The smoothing is applied recursively for lower-order n-grams.

Headlines were extracted from the MN-DS news corpus and provided a corpus of real-world headlines to train the model. To ensure training integrity, the MN-DS dataset was filtered to include only headlines with more than eight tokens, based on the premise that longer headlines provide richer contextual information crucial for effective language modeling. N-gram counts were calculated, focusing on unigrams, bigrams, and trigrams. Each headline was tokenized into lowercase words using the NLTK library's word tokenizer. Counts were aggregated into three distinct dictionaries: unigram_counts, bigram_counts, and trigram_counts, serving as foundations for estimating n-gram probabilities in subsequent perplexity calculations. Perplexity was computed for each headline in both datasets using trigrams extracted from tokenized headlines. Kneser-Ney smoothing was applied to address challenges posed by unseen n-grams, with smoothing parameters set to $\alpha = 0.75$ and $\gamma = 0.5$. Perplexity for each headline was derived from the exponential of the average negative log probability of its constituent trigrams. Undefined perplexity values were replaced with NaN to maintain dataset integrity and facilitate subsequent statistical analyses.

### 3.4.3. Readability

Furthermore, we undertake comparative style and readability analysis using the Flesch-Kincaid readability tests [27]. The metrics assess the complexity of the text by considering factors such as sentence length and word syllable count to determine the ease with which a reader can comprehend the headlines. If the generated headlines exhibit readability scores that fall within the range typical for mainstream news outlets, it may signify adherence to journalistic standards of clarity and conciseness.

The score is computed as shown in Equation 3.

$$Readability = 0.39 \cdot \left(\frac{\text{total words}}{\text{total sentences}}\right) + 11.8 \cdot \left(\frac{\text{total syllables}}{\text{total words}}\right) - 15.59 \qquad (3)$$

The Flesch-Kincaid Grade Level, a widely adopted metric, provides an estimate of the US grade level required to comprehend a given text. The computational implementation of this readability measure, focusing on a robust approach utilizing the Carnegie Mellon University Pronouncing Dictionary (CMUdict) was used for enhanced accuracy. To implement this calculation, the process is divided into distinct steps:

- Text Preprocessing: The input text undergoes tokenization to identify sentences and words using the nltk.tokenize module.
- Syllable Counting: While simpler methods estimate syllables based on word length, this implementation leverages the CMUdict for greater precision. This method provides access to a comprehensive pronunciation dictionary. For each word in the text, the dictionary is queried to retrieve its phonetic representation. Syllables are then counted by identifying vowel sounds within the phonetic transcription. In cases where a word is absent from the dictionary, a fallback approximation based on word length is employed.
- Formula Application: With the total number of words, sentences, and syllables determined, the Flesch-Kincaid Grade Level is calculated using Equation 3.

### 3.4.4. BERTScore

We employed semantic similarity measures using BERTScore, a metric that leverages contextual embeddings from transformer-based models [28]. BERTScore aligns each token in a generated headline with tokens in reference headlines based on semantic meaning instead of exact word matches. This method is particularly advantageous in this case since it captures the semantics and thematic relevance of the headlines. The BERTScore metric is used to evaluate the semantic similarity between automatically generated headlines and headlines from a manually curated dataset (MN-DS) of human news. Using pre-trained contextual embeddings from the `bert-base-uncased` model, we achieve a better assessment of semantic similarity compared to traditional methods that rely solely on word overlap. This satisfies the need to capture semantic equivalence rather than mere lexical similarity, recognizing that paraphrases or synonyms could convey the same meaning while having different wordings. The average BERTScore of the headlines would indicate a high, or low degree of semantic similarity between our generated headlines and a reference set of authentic news headline, thereby affirming if the content is contextually appropriate and mirrors real-world news scenarios or not.

### 3.4.5. Semantic similarity

Furthermore, we conducted comparative semantic similarity analysis to assess the thematic relevance and contextual appropriateness of the generated negative news headlines. The analysis was undertaken using sentence transformers and by using pre-trained embeddings.

a. Sentence transformer

This analysis employed a SentenceTransformer model, precisely the 'all-mpnet-base-v2', which leverages advanced transformer architecture to generate high-quality sentence embeddings [29]. The use of such contextual embeddings allows for a comparison that goes beyond surface-level lexical similarities, capturing deeper semantic relationships between headlines. We created embeddings for the reference headlines obtained from Newsnow.com, which served as our benchmark for authentic news content. We generated embeddings for the synthetic negative news headlines, and computed cosine similarity scores between each generated headline and the entire set of reference headlines. Cosine similarity is an established measure for examining how closely related terms are in NLP [30]. The maximum similarity score for each headline was retained, providing a measure of how closely it aligned with the most semantically similar reference headline. To gain a more comprehensive understanding of the semantic similarity scores, we conducted a comparative analysis utilizing MN-DS, a dataset of 10,821 authentic news headlines [25]. To ensure a valid comparison for some parts of the analysis, we restricted our analysis to headlines comprising eight words or fewer, in line with the average word count in our synthetic dataset. This constraint serves to mitigate potential biases arising from variations in headline length and complexity.

b. GloVe word embedding

To further quantify semantic similarity within text data, we also leverage the GloVe model. Word embedding models, such as GloVe, have been shown to effectively capture semantic relationships between words [31]. GloVe was pre-trained on the Wikipedia and Gigaword corpus. This approach allowed us to analyze the semantic relationship between words in the headlines. GloVe is one of the various ways of examining semantic meaning in text and one of the most widely used pre-trained word embeddings [30]. Our methodology involved tokenizing sentences into individual words, converting them to lowercase, and then computing the cosine similarity between each word pair using the GloVe model. The similarity scores were subsequently averaged to obtain a single similarity value per sentence. By applying this methodology to our datasets, we aimed to uncover insights into the semantic relationships between words in text headlines. The datasets used in the sentence transformer analysis were similarly used in the same approach.

### 3.4.6. Part of Speech (POS) profile

Furthermore, we employed a deeper approach to evaluate the stylometric characteristics of two sets of headlines: the synthetic dataset and a reference set of 66 real time news headlines obtained from Newsnow.com at a point during this research. This reference dataset served as our comparison benchmark. We initiated our evaluation by examining the Part-of-Speech (POS) distributions within both datasets. Utilizing the Stanza pipeline, which is based on CoreNLP, we performed POS tagging on the preprocessed headlines. The preprocessing phase involved the removal of stopwords and punctuation, leaving the core lexical elements of each headline. We processed the headlines NLP techniques to tokenize and tag parts of speech. We then calculated POS ratios for each headline, allowing for a normalized comparison across datasets of different sizes.

Given a set of tokens $(T = t_1, t_2, \ldots, t_n)$ in the news headline, the Part-of-Speech (POS) tags are extracted as:

$$POS(T) = \{ pos(t_i) \mid t_i \in T \} \tag{4}$$

For a list of POS tags $(\mathrm{POS(T)})$ the frequency $(f_t)$ of a specific tag is given by:

$$f_t = \sum_{i=1}^{n} 1(t_i = t) \tag{5}$$

where $\mathbb{1}(t_i = t)$ *is* the indicator function that equals 1 if $t_i = t$ and *0* otherwise.

For a document $D$ with a total of $N$ tags, the ratio $r_t$ of tag $t$ is:

$$r_t = \frac{f_t}{\sum_{t' \in T} f_{t'}} \tag{6}$$

Where $f_t$ is the frequency of tag $t$ in $D$.

Statistical analysis was conducted using independent t-tests to compare these ratios between the two datasets, providing a rigorous assessment of the significance of observed differences.

## 4.0. Benchmarking results

Based on the corpus level evaluation across multiple dimensions, we examine the results from the various aspects of the tests. To benchmark the results from the synthetic dataset, we employed the MN-DS dataset and a subset of it, and a collection of real time news items from Newsnow.com. Details about the dataset are presented in Table 1. Figure 10 shows a set of visualizations of the various datasets mainly with respect to their tokens. Figure 11 shows a box plot comparing the token length of the three.

Table 1: Datasets used in the evaluation and their characteristics.

| Dataset | Number of headlines | Token length per headline |
|---|---|---|
| MN-DS Short | 976 | <=8 |
| MN-DS | 10821 | <=45 |
| Real News | 66 | <=18 |
| Synthetic News | 510 | <=10 |

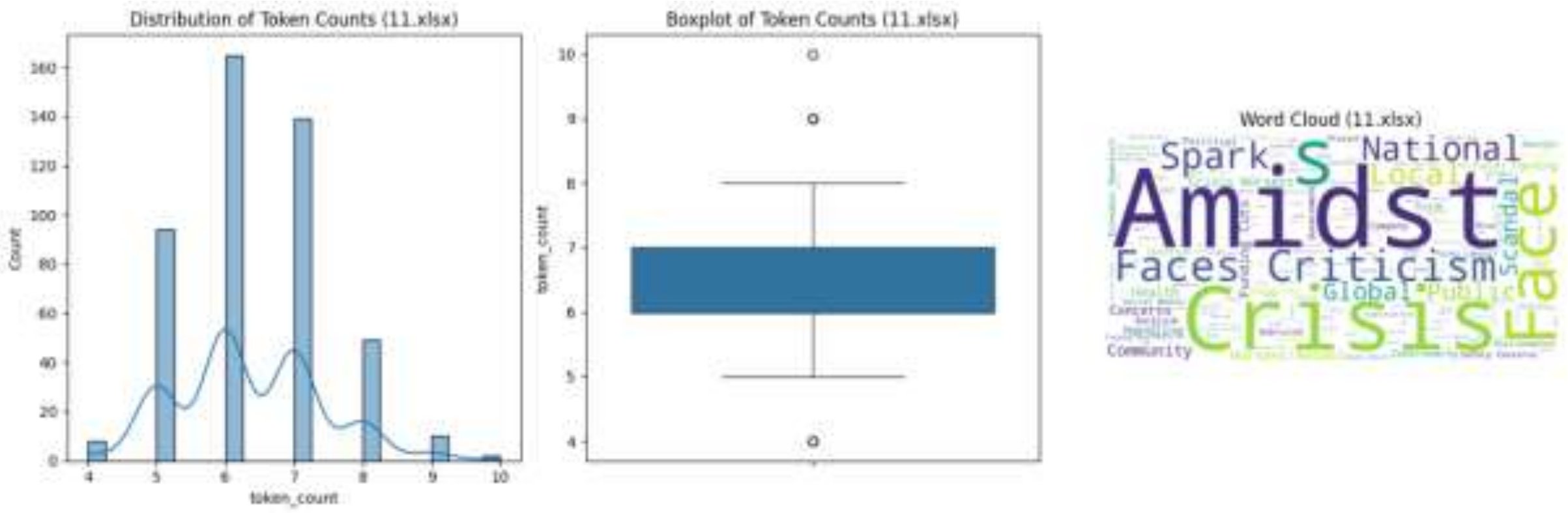

Figure 10a. Distribution of token length, box plot of token length and a wordcloud of the generated dataset

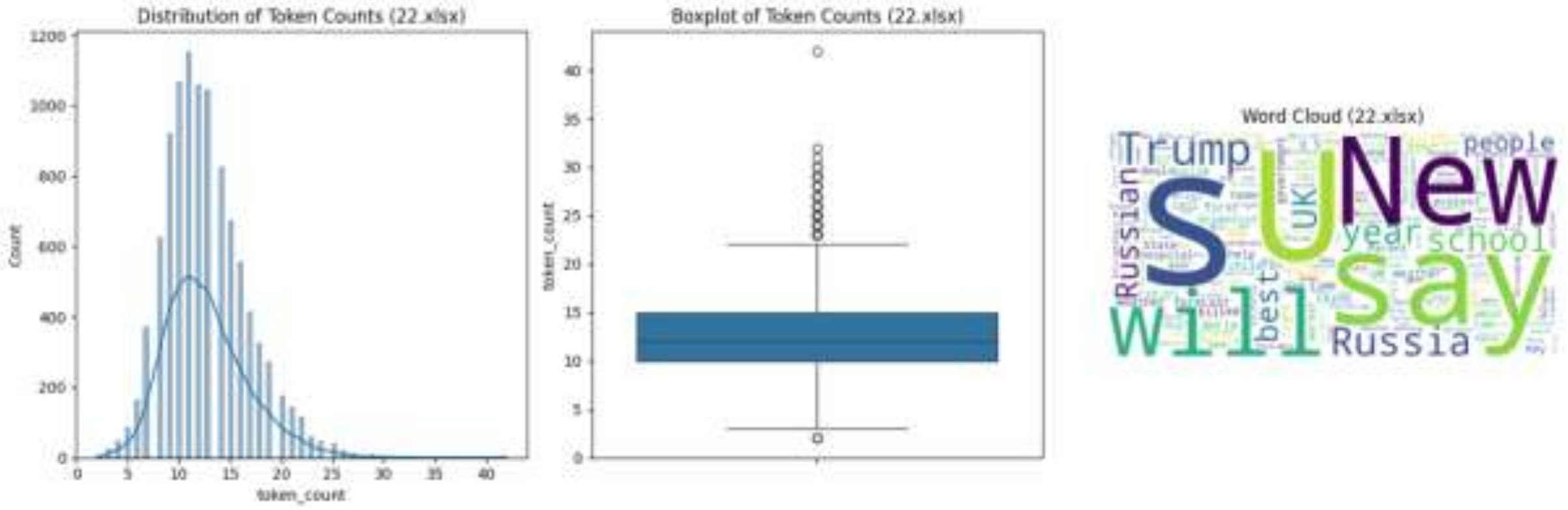

Figure 10b. Distribution of token length, box plot of token length and a wordcloud of the MN-DS news headlines dataset.

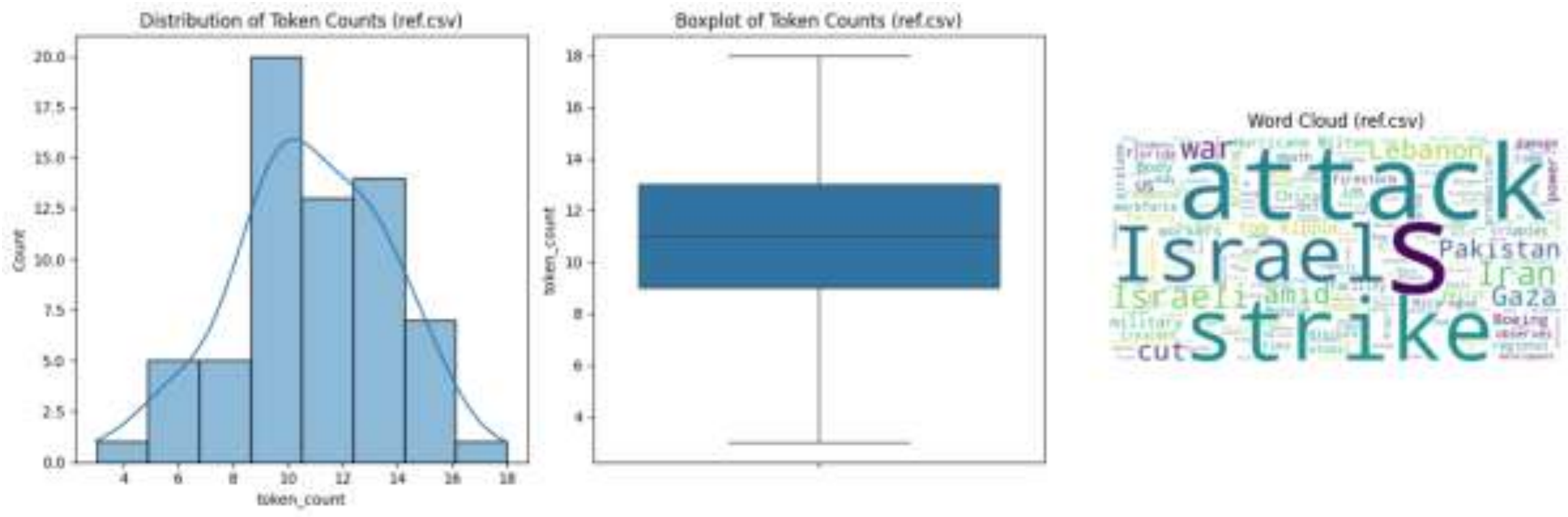

Figure 10c. Distribution of token length, box plot of token length and a wordcloud of the reference realtime news dataset.

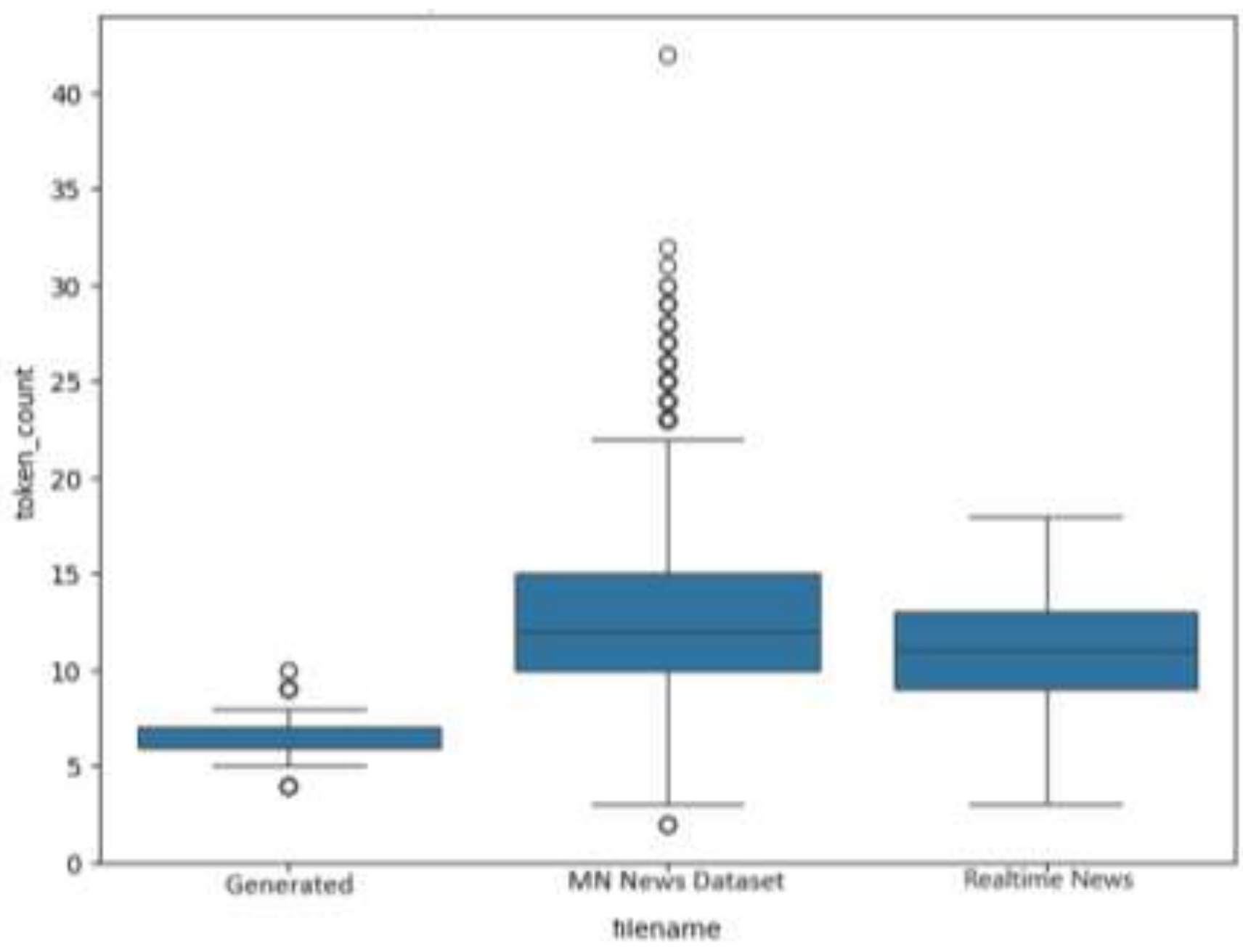

Figure 11. Box plot of token counts for the three datasets involved in the various evaluations.

## 4.1. POS profile evaluation

The statistical analysis shows significant differences in part-of-speech (POS) ratios between main (synthetic news dataset) and reference headlines. Specifically, notable variations in the ratios of nouns, pronouns, adpositions, proper nouns, and adjectives. The charts for POS profiling are shown in Figure 12.

Notably, the use of proper nouns (PROPN) exhibited the most pronounced disparity (*t-statistic = -2.1885, p-value = 0.0290*). This extremely low p-value indicates an overwhelming statistical significance, suggesting that reference headlines incorporate substantially more proper nouns than the main dataset. The charts in Figure 12 and the POS wordcloud in Figure 13 illustrate and support this observation. This result may indicate that headlines in the synthetic news dataset tend to avoid using pronouns, which can create a sense of distance or objectivity, and a greater emphasis on specific entities, individuals, or locations in the reference headlines.

Table 2a. Normalized POS frequency result for the datasets (in %).

| POS | Synthetic dataset (Main) | Reference dataset (Realtime news) |
|---|---|---|
| VERB | 17.7% | 16.9% |
| PRON | 0.1% | 0.6% |
| SCONJ | 0.0% | 0.0% |
| X | 0.0% | 0.2% |
| ADP | 4.8% | 1.3% |
| PROPN | 0.5% | 15.4% |
| ADV | 0.8% | 0.6% |
| NOUN | 61.3% | 51.3% |
| AUX | 0.0% | 0.6% |
| ADJ | 14.9% | 10.2% |
| NUM | 0.0% | 2.9% |

Table 2b. POS distribution of the main and reference datasets (actual numbers).

| POS | Main | Reference |
|---|---|---|
| VERB | 507 | 88 |
| PRON | 2 | 3 |
| SCONJ | 1 | 0 |
| X | 0 | 1 |
| ADP | 138 | 7 |
| PROPN | 13 | 80 |
| ADV | 22 | 3 |
| NOUN | 1755 | 267 |
| AUX | 0 | 3 |
| ADJ | 426 | 53 |
| NUM | 0 | 15 |

Table 3. Statistical analysis of the POS analysis results.

| POS | t_statistic | p_value |
|---|---|---|
| VERB_Ratio | 0.9022 | 0.3673 |
| NOUN_Ratio | 5.6730 | <0.0001 |
| PRON_Ratio | -2.1885 | 0.0290 |
| ADP_Ratio | 3.4454 | 0.0006 |
| PROPN_Ratio | -21.2579 | <0.0001 |
| ADV_Ratio | 0.4254 | 0.6707 |
| ADJ_Ratio | 2.9943 | 0.0029 |

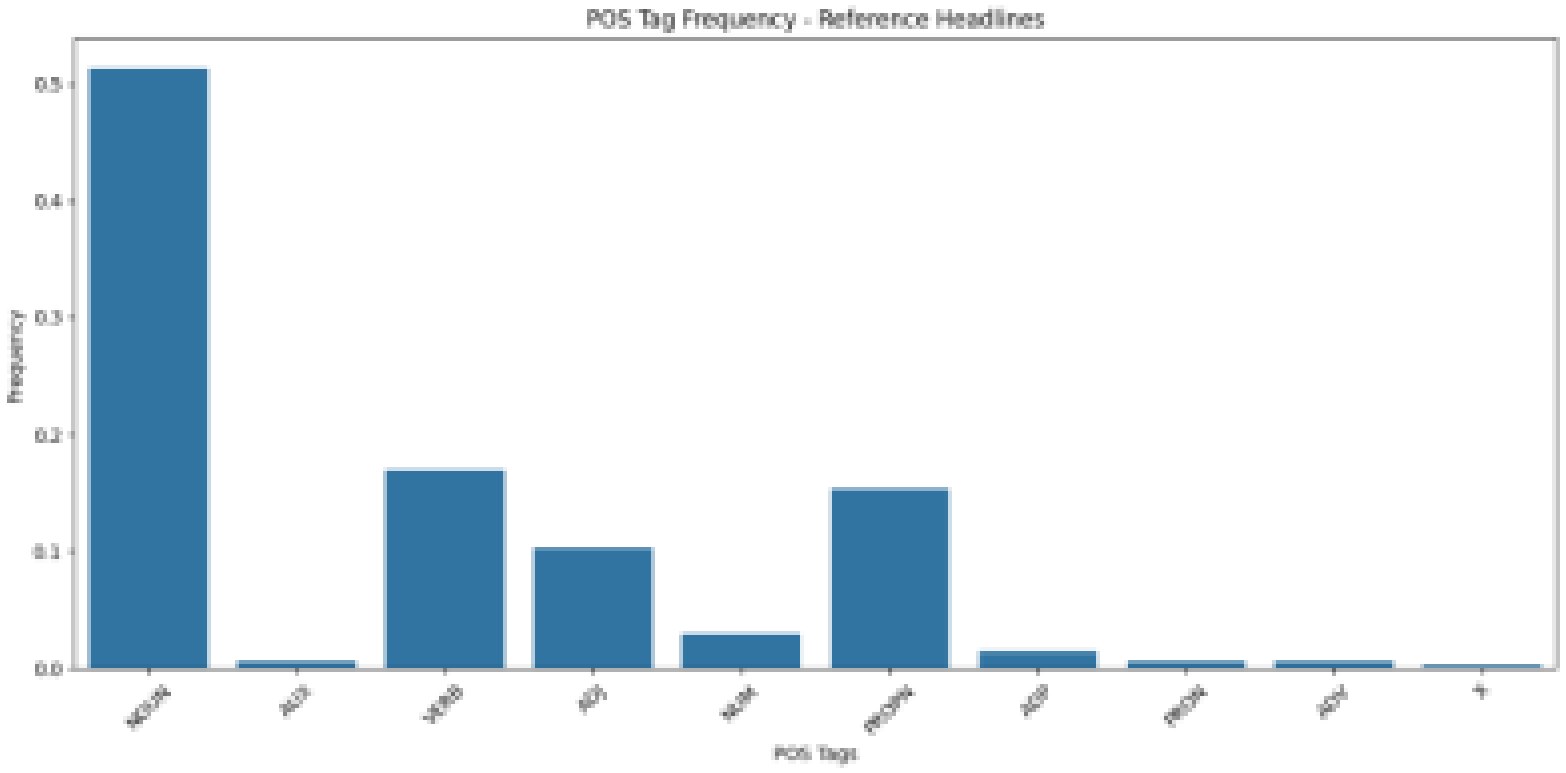

Figure 12a.

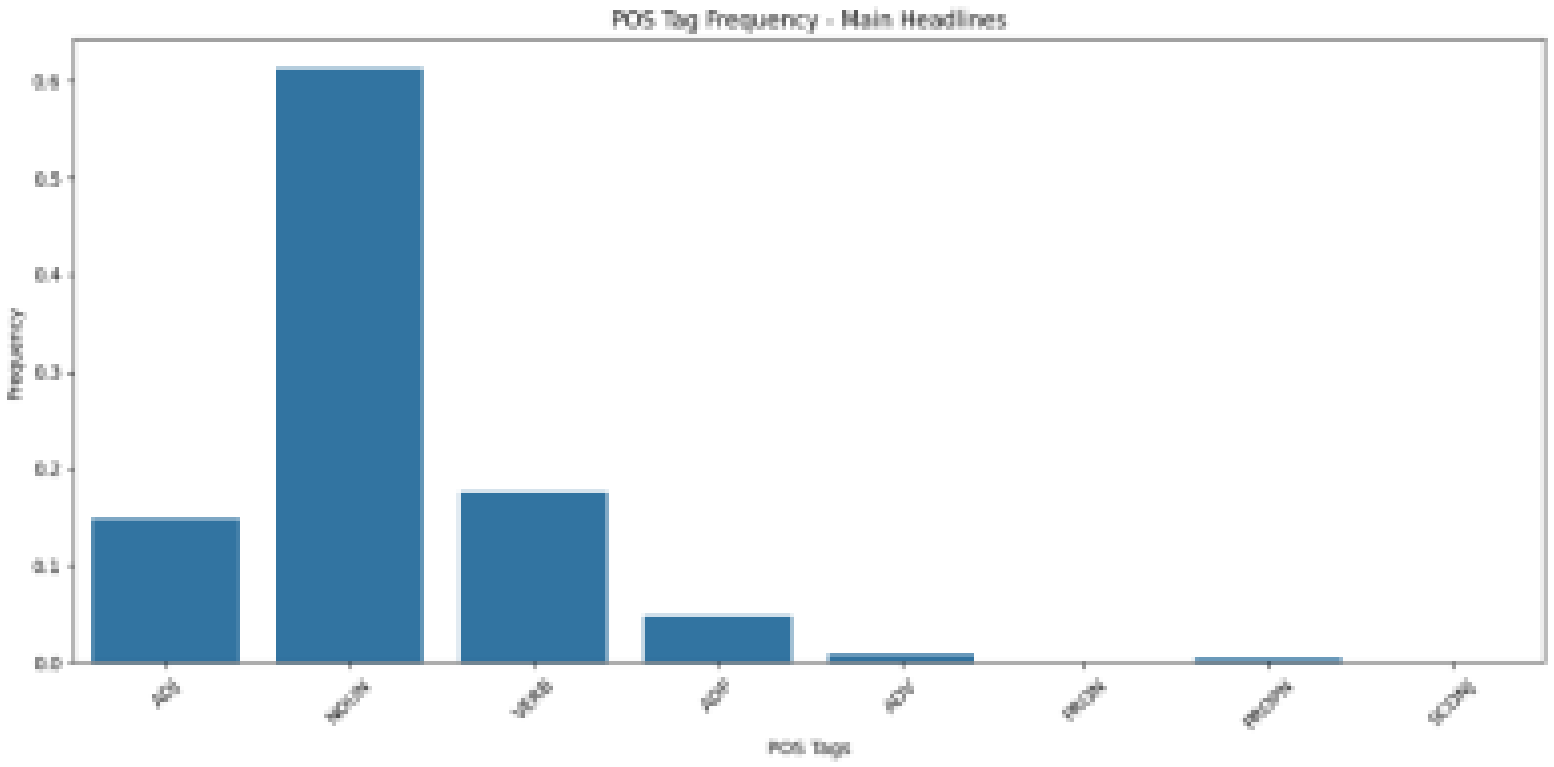

Figure 12b.

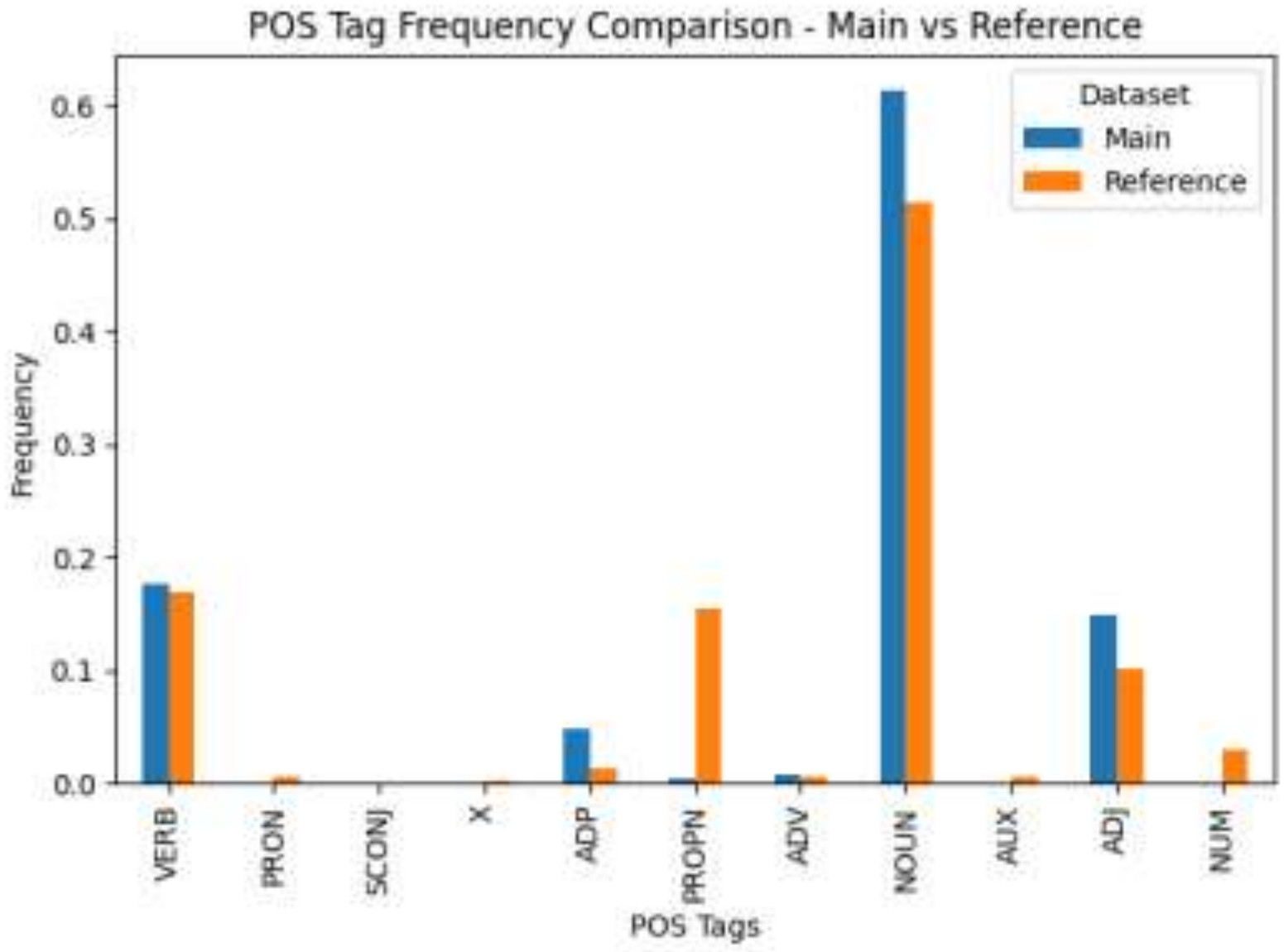

Figure 12c.

Figure 12. Normalized part of speech distribution for the reference corpus, the synthetic news corpus (main), and the combined chart of both.

The analysis showed that the ratio of nouns was significantly higher in main headlines compared to reference headlines (*t-statistic = 5.6730, p-value < 0.0001*). This finding suggests that main headlines tend to employ more nominal phrases, which may be indicative of a more formal or objective tone. It could also indicate a more generalized or abstract approach to topic presentation.

Adpositions (ADP) and adjectives (ADJ) were more prevalent in the main dataset (*t-statistics: 3.4454* and *2.9943, p-values: 0.0006 and 0.0029*, respectively), potentially signifying a tendency towards more detailed or descriptive content as they play a crucial role in establishing relationships between entities in a sentence. Conversely, pronouns (PRON) were more frequently employed in reference headlines (*t-statistic: -2.1885, p-value: 0.0290*), which might suggest a more personalized or narrative style in these headlines.

Interestingly, the analysis revealed no statistically significant differences in the usage of verbs (VERB) and adverbs (ADV) between the two datasets (*p-values: 0.3673* and *0.6707*, respectively). This similarity in verb and adverb usage implies a certain consistency in action description and modification across both sets of headlines. These findings illuminate the fine differences in grammatical structure and lexical choice between the main and reference headlines. The variations in POS ratios between main and reference headlines suggest that different linguistic strategies are employed in these two contexts.

The divergence in proper noun usage is particularly noteworthy, as it may reflect distinct strategies in the main content idea and presentation. The higher incidence of common nouns, adpositions, and adjectives in the main dataset suggests a potential inclination towards more generalized yet descriptive headline construction.

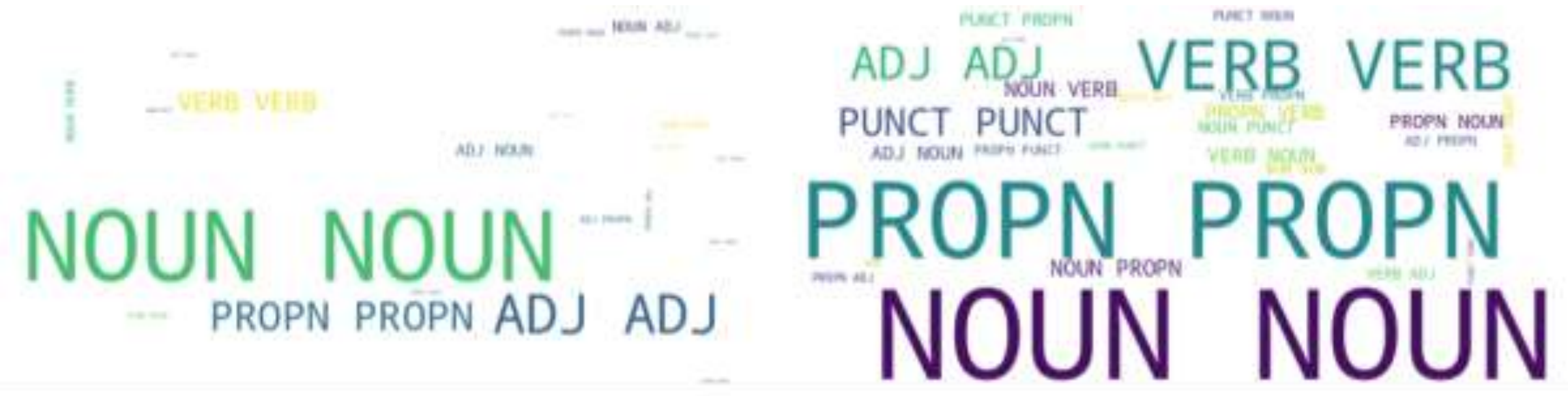

Figure 13: Wordcloud of POS distribution in synthetic negative news dataset (left), and in real news dataset (right)

## 4.2. Semantic similarity evaluation

This study investigated the distribution of semantic similarity scores for real news headlines using sentence transformers and pre-trained word embeddings. The resulting similarity scores from our testing provide a quantitative measure of how well the generated headlines mirror the semantic content typical of real-world news scenarios. Higher scores indicate a stronger alignment with authentic news headlines, suggesting that the generated content is contextually appropriate. Conversely, lower scores might indicate divergence from typical news content, potentially highlighting areas where the headline generation process could be refined.

a. Semantic evaluation using sentence transformer

We investigated the semantic similarity between news headlines from three distinct datasets: MN-DS short (976 headlines), MN-DS (10,821 headlines), and synthetic negative news (467 headlines) using SentenceTransformer model (all-mpnet-base-v2). We analyzed two sets of real news headlines: a smaller set of 976 headlines and a larger set of 10,821 headlines. We also examined a synthetic negative news dataset containing 510 headlines. Specifically, we calculated the mean semantic similarity scores for each dataset. The results showed that the synthetic dataset headlines exhibited the highest mean semantic similarity (0.2809), closely followed by MN-DS (0.2770) and MN-DS short (0.2759). This suggests that the generated headlines are more semantically like the reference headlines. When examining the mean similarity scores, a small but significant difference was observed. This slight increase in mean similarity for the synthetic headlines could indicate that the generation model successfully captured common semantic patterns found in news reporting. However, caution is required in interpreting this finding, as the difference is small and may not necessarily mean a noticeable difference in the quality or authenticity of the headlines.

To further understand the dispersion of semantic similarity scores, we computed the standard deviation for each dataset. The synthetic dataset headlines demonstrated the lowest standard deviation (0.0694), indicating more consistent semantic similarity scores. In contrast, MN-DS short and MN-DS showed slightly higher standard deviations (0.0868 and 0.0839, respectively). The high semantic similarity between the synthetic dataset headlines and the reference headlines compared to the others implies that these headlines share common semantic patterns. Moreover, the comparable semantic similarity between MN-DS and MN-DS short headlines should be expected since they are the same dataset split by headline length. The difference in standard deviation suggests that human-produced news headlines display a wider range of semantic variation compared to the synthetic headlines. This can be interpreted in different ways; the variation that may be due to the diverse nature of real-world events and the different ways journalists write headlines, it may be due to the monotonous valence particular to the synthetic dataset, or simply due to the shorter length of the headlines and the small size of the corpus.

Table 4. Semantic similarity evaluation data using sentence transformers.

| Dataset | Number of headlines | Standard Deviation | Mean |
|---|---|---|---|
| Real news (MN-DS short) | 976 | 0.086776 | 0.275986 |
| Real news (MN-DS) | 10821 | 0.083882 | 0.277017 |
| Synthetic News | 467 | 0.06938 | 0.280887 |

b. Semantic evaluation using GloVe embeddings

Furthermore, we evaluated similarity across the three distinct datasets: Generated, realtime news, and MN-DS Short using GloVE. Our analysis revealed notable differences in similarity metrics across the three datasets. The synthetic dataset exhibited the highest mean similarity (0.434), followed by the Realtime dataset (0.4045), and then the MN-DS Short dataset (0.3773). A similar trend was observed for median similarity, with the synthetic dataset having the highest value (0.4398). The standard deviation of similarity, which measures the spread of similarity values, was lowest for the synthetic dataset (0.0851) and highest for the MN-DS Short dataset (0.1315). This suggests that the synthetic dataset has a more consistent level of similarity across its entries, while the MN-DS Short dataset has a more variable level of similarity. Generated headlines displayed the lowest standard deviation indicating consistent semantic similarity, in contrast, MN-DS Short headlines exhibited the highest standard deviation (0.1315), suggesting greater semantic variability.

This suggests that generated headlines are again semantically more aligned with the GloVe embeddings, while realtime headlines show a moderate connection. The median similarity scores reinforced this notion, with generated headlines scoring 0.4398, realtime headlines scoring 0.4231, and MN-DS Short headlines scoring 0.3832.

Our results indicate that the GloVe similarity evaluation for the synthetic dataset had the best mean semantic similarity scores showing it is characterized by a higher degree of semantic coherence. In contrast, the realtime dataset, which is comprised of real-world text data, exhibits a lower mean similarity. This may be attributed to the inherent noise and variability as these were just the latest news headlines from all over the world at that point in time. The MN-DS Short dataset, which is the set of news headlines from the MN-DS dataset with token length <= 8, exhibits the lowest mean semantic similarity score. This disparity contrasts with the synthetic dataset which though it has short length has better semantic scores.

Table 5. Similarity evaluation data using GloVe embeddings.

| Dataset | Mean Similarity | Median Similarity | Standard Deviation of Similarity |
|---|---|---|---|
| Generated | 0.4340 | 0.4398 | 0.0851 |
| Realtime | 0.4045 | 0.4231 | 0.1141 |
| MN-DS Short | 0.3773 | 0.3832 | 0.1315 |

## 4.3 Perplexity evaluation

Our perplexity analysis examined the language patterns of three datasets: Generated, Realtime, and MN-DS Short. By employing a Kneser-Ney smoothed n-gram model, we uncovered significant differences in the average perplexity scores between the datasets, the summary result is presented in Table 6 and Figure 14. Notably, MN-DS Short headlines exhibited the highest perplexity score of 1.5700, indicating greater uncertainty or unpredictability in their language patterns. In contrast, generated headlines showed the lowest perplexity score of 1.1557, suggesting more predictable language. Realtime headlines fell between the two, with a perplexity score of 1.2184, indicating moderate language complexity. The high perplexity score of MN-DS Short headlines suggests that they are more diverse or complex. Conversely, the low perplexity score of generated headlines indicates that they demonstrate more predictable language patterns.

Furthermore, our category-wise perplexity analysis of generated headlines revealed varying levels of perplexity (see Figure 15). By normalizing perplexity by category count, we gained insights into language patterns within each category. Conventionally, lower perplexity scores are associated with more predictable and fluent text. In this case, the lower perplexity of the synthetic headlines suggests that they adhere more closely to common linguistic patterns found in news reporting. This could indicate that the generation model has effectively captured and reproduced typical headline structures and vocabulary usage. However, it is crucial to note that excessively low perplexity might also signify a lack of linguistic diversity or creativity. Human-written headlines, with their higher perplexity score, may reflect a greater variety in language use, potentially incorporating more novel or unexpected phrasings that deviate from the most common patterns.

Table 6. Perplexity results data.

| Dataset | Average Perplexity |
|---|---|
| Generated | 1.1557 |
| Realtime | 1.2184 |
| MN-DS Short | 1.5700 |

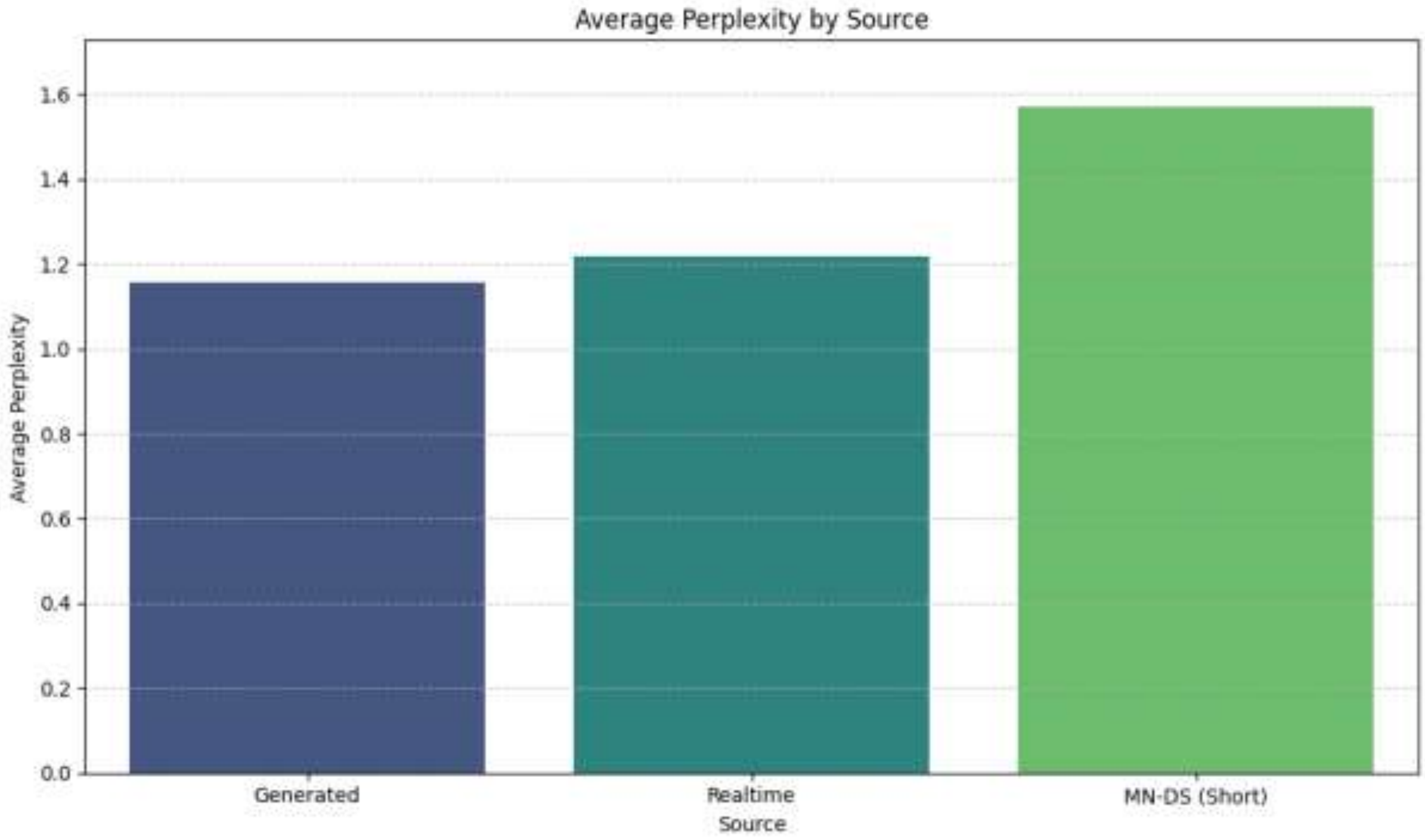

Figure 14. Average perplexity distribution in the evaluated datasets.

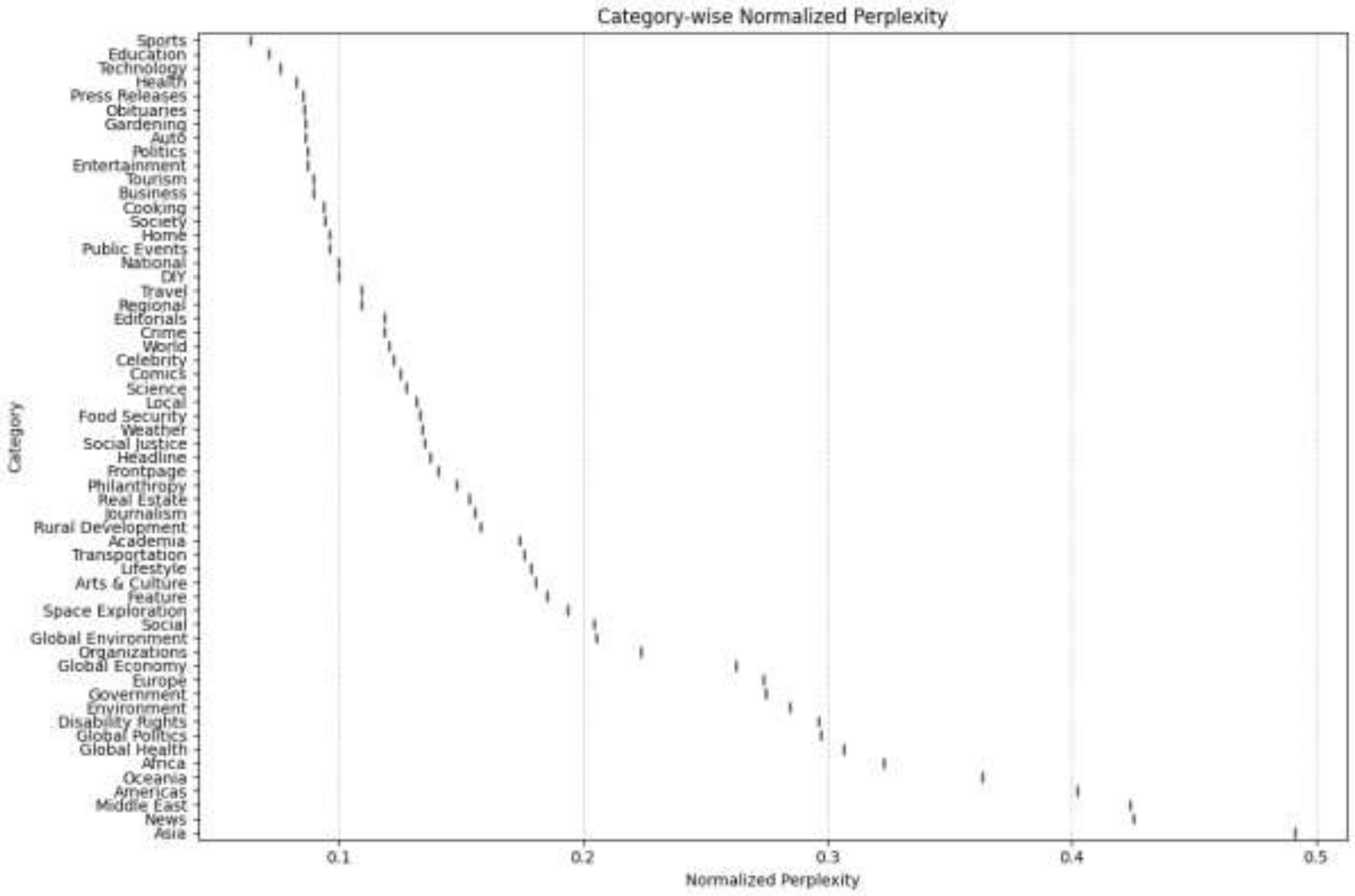

Figure 15. Perplexity distribution by category in the generated dataset.

## 4.4 Readability evaluation

For the readability test, generated headlines exhibited the highest average Flesch-Kincaid score of 12.78, indicating a comparatively higher grade level of readability. In contrast, realtime headlines demonstrated the lowest average Flesch-Kincaid score of 9.65, suggesting a simpler presentation language is found in this dataset. MN-DS Short headlines fell between the two, with an average Flesch-Kincaid score of 9.8 as seen in Table 7 and Figure 16. These findings imply that generated headlines are more complex compared to Realtime and MN-DS Short headlines. Interestingly the headlines in the Realtime dataset have headlines with longer token length than those in the generated dataset or the MN-DS Short dataset. The shorter length of the synthetic headlines notwithstanding, it exhibited the highest readability complexity.

Further analysis of generated headlines by category revealed varying levels of readability. The variability in the generated dataset's readability is across the categories and is pronounced. While some categories had readability of 6.2, others had values as high as 13.8 as can be seen in Figure 18.

Table 7. Flesch-Kincaid results.

| Dataset | Average Flesch-Kincaid |
|---|---|
| Generated | 12.78 |
| Realtime | 9.65 |
| MN-DS Short | 9.8 |

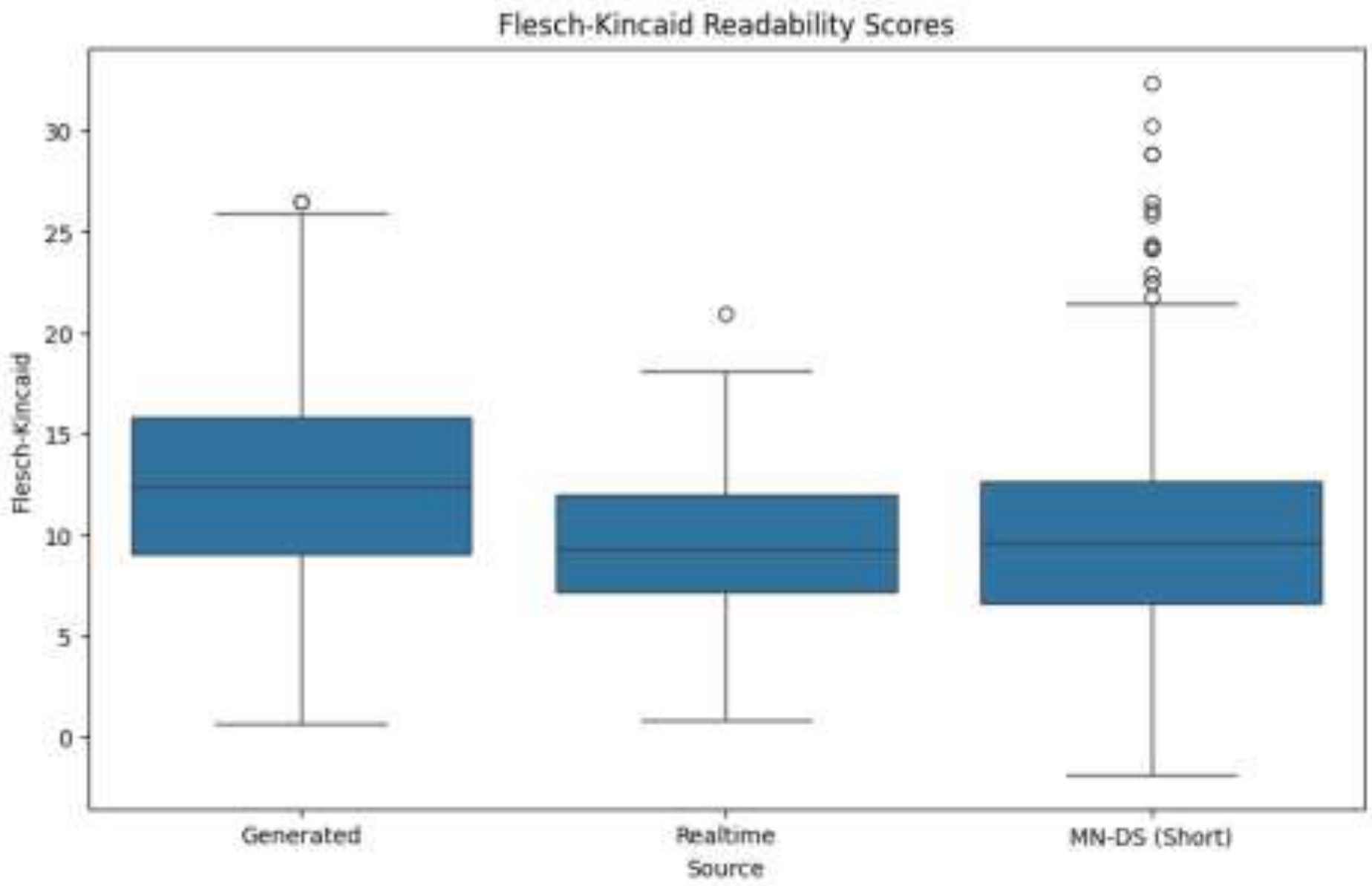

Figure 17. Comparative box plot of readability scores.

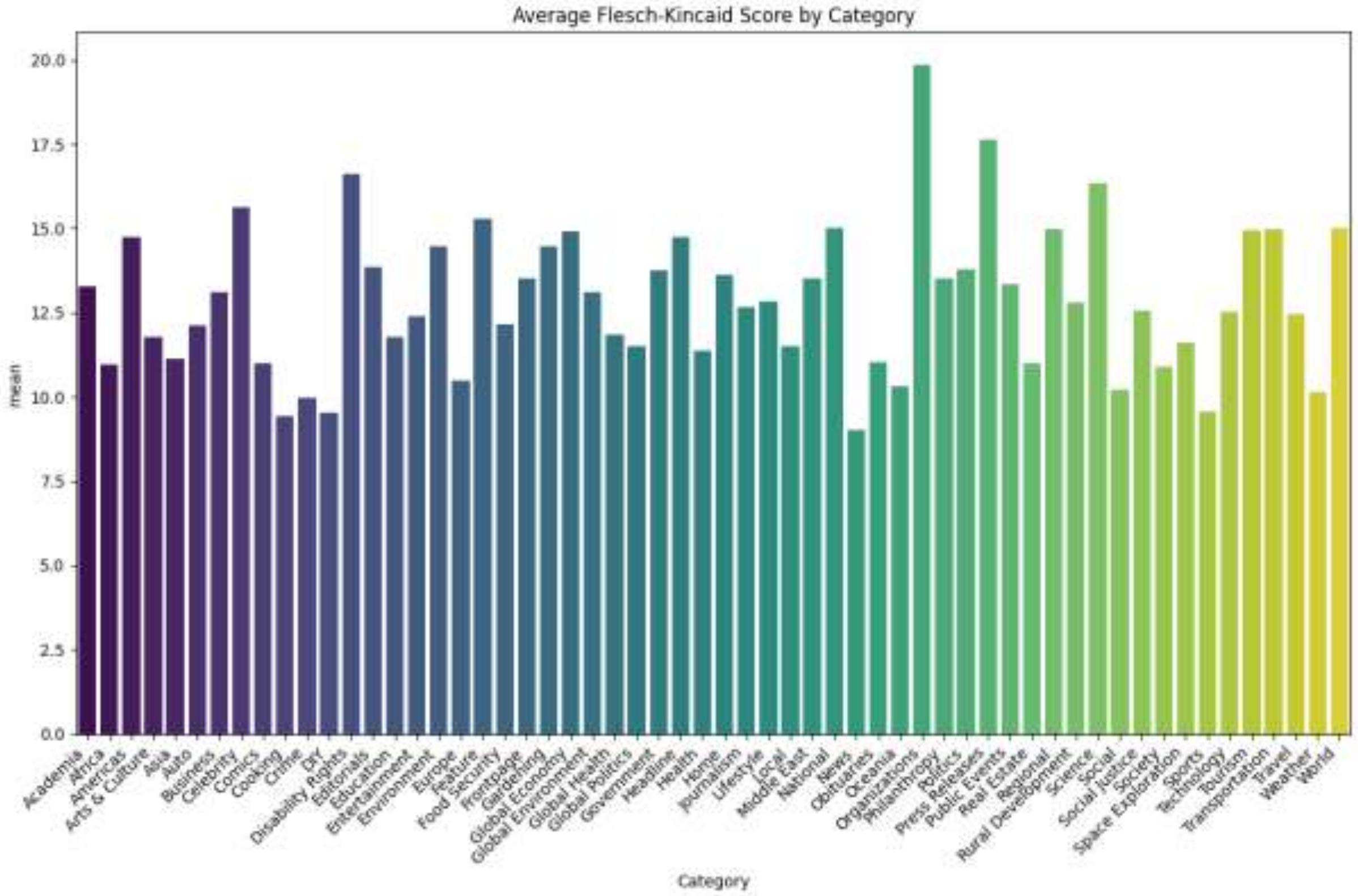

Figure 18a. Chart of readability score per category.

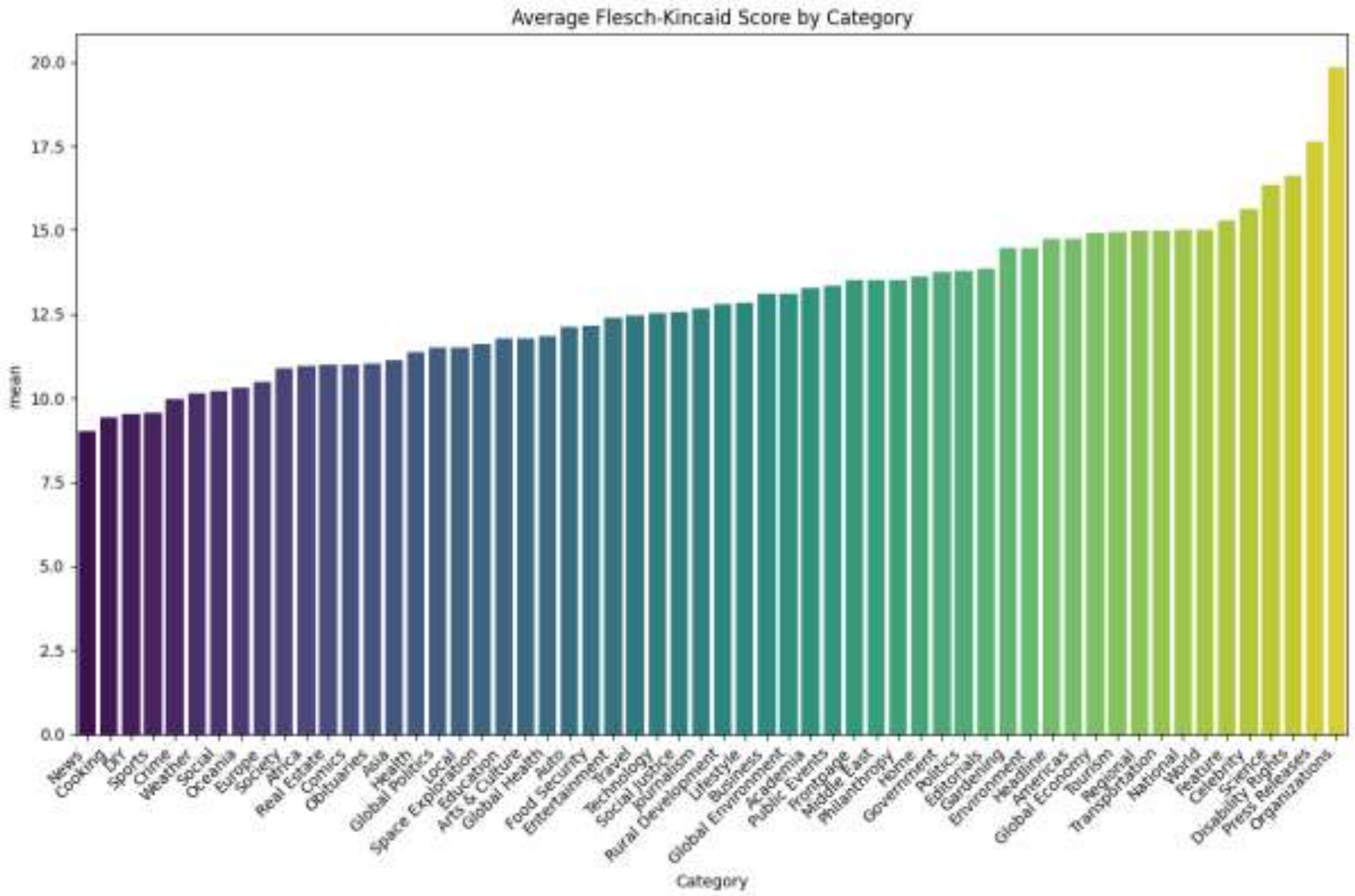

Figure 18b. Score-ordered chart of readability score of the synthetic dataset.

## 4.5 BERTScore evaluation

The methodology involved a pairwise comparison of generated headlines with headlines from the MN-DS dataset. Each generated headline was compared to a randomly selected subset of `X` unique headlines from the MN-DS. This sampling strategy helped maximize diversity in the comparison set as well as providing a robust measure of similarity in a more representative manner. In all we compared the highest F1 scores for a portion of the generated dataset (302 headlines) and highest F1 scores for the realtime dataset (64 headlines) using statistical tests designed for independent samples as these tests do not require paired data and can manage missing values more effectively.

Table 8. Descriptive statistics for BERTScore evaluation results.

| Statistic | Generated Headlines | Realtime Headlines |
|---|---|---|
| count | 302 | 64 |
| mean | 0.508748 | 0.489105 |
| std | 0.035839 | 0.038587 |
| min | 0.43679 | 0.362979 |
| 25% | 0.484111 | 0.462845 |
| 50% | 0.503971 | 0.490988 |
| 75% | 0.530672 | 0.51827 |
| max | 0.663184 | 0.59195 |

The statistical test yielded significant results, as summarized in Table 9.

Table 9. T-test statistics for BERTScore results for generated and realtime headlines

| Statistic | Value |
|---|---|
| T-statistic | 3.9291 |
| P-value | 0.0001 |

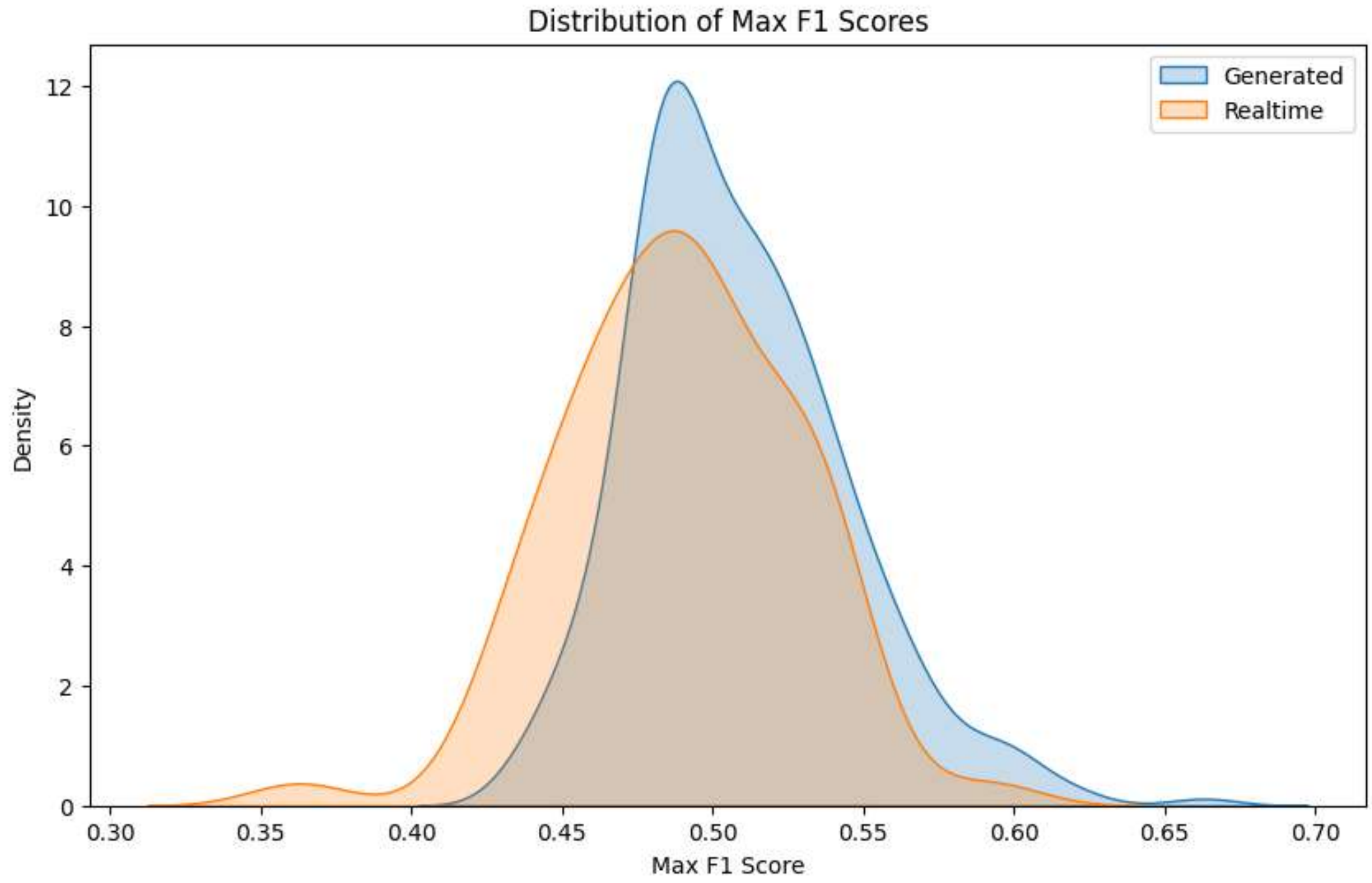

Figure 19. Density plot of F1 scores distribution.

The T-statistic and P-value provide convincing evidence that the automatically generated headlines outperform the real-time dataset with respect to semantic similarity to human produced news headlines. The results indicate that the generated headlines exhibit a statistically significant higher average F1 score compared to the real-time dataset. This suggests a promising performance of the LLM in producing semantically similar headlines to the MN-DS news dataset, as assessed by BERTScore.

# 5.0. Discussion and conclusion

The study both generated and evaluated a dataset of negative news. The focus was on benchmarking the realism of the generated news headlines.

## 5.1. Discussion

POS profiles between synthetic negative news headlines generated by LLM and human-produced news headlines have major differences, especially in the use of proper nouns where synthetic news rarely had instances of proper nouns. Furthermore, the profiles also differ for adpositions where human-produced news rarely used adpositions. Noticeable differences also exist in the use of nouns and adjectives, with more of them found in the synthetic headlines.

The generated dataset had a lower perplexity score than the dataset of human-generated news headlines. Both results favor the synthetic dataset. However, the Flesch-Kincaid score for the synthetic dataset was about 25% higher than for the human generated news. However, evaluating the semantic composition of the datasets showed the generated dataset had a slightly higher semantic score in both the GloVe and sentence transformer evaluations. Over these multiple examinations, the LLM-generated news headlines can pass as real news headlines except in one noticeable area, their part of speech composition.

Our evaluation of generated negative news headlines raises a question whether the headlines demonstrate fluency and coherence but lack diversity and semantic complexity compared to human-written news. Reasons for this shortcoming include the fact that the dataset is a negative news dataset being compared with a dataset of general news, or the fact that there is a POS deficiency (proper noun) in LLM generated news.

The results clearly show the LLM generated news as being devoid of proper nouns, this glaring absence suggests intentionality on the part of the LLM. It is safe guess that putting a proper noun in a synthetic negative news would be unacceptable. But without proper nouns, there would always be a marked difference between a synthetic corpus and a real news corpus. The results underscore at least one major limitation of current LLMs in totally replicating the model and variability of human language, particularly in domain-specific contexts such as news writing. This warrants consideration when using public language models for generating text datasets.

## 5.2. Conclusion

By leveraging the LLMs-Driven Synthetic Data Generation, Curation, and Evaluation framework, the study provides a robust methodology for generating, curating, and evaluating synthetic news headline datasets using an LLM. This methodology provides a systematic approach to generating and validating a synthetic dataset of news headlines using an LLM as a tool. The dataset's focus on negative sentiment across multiple news categories allows for a comprehensive evaluation of sentiment analysis tools in production environments.

This approach combines the strengths of LLM-driven data generation with curated feedback loops to refine the final dataset. By adopting this framework, this research ensures the generation process produces a high-quality, domain-relevant dataset that reflects real-world linguistic patterns while maintaining controlled sentiment distributions. This study benefits from LLM's advanced capabilities to generate headlines across 58 categories, with prompts designed to yield negative news. The model's contextual understanding ensures that generated headlines are not just syntactically correct but also reflect the sentiment nuances critical to this study. The results from this study, although with a small-sized dataset yield some interesting knowledge about negative news and the use of LLMs for generating synthetic negative news headline data.

Regarding the dataset generated, the coherence of the generated news headlines was perfect. The headlines match the journalistic style found in most standard news outlets. The LLM was able to generate headlines for each category of news in various domains, all headlines were well captured with diverse angles, coverages, and stories. Overall, LLM produced an engaging set of headlines. The sentiments were all real in the sense that all the issues mentioned in the headlines were plausible; the headlines generated exhibited both engagement and plausibility.

While our study provides valuable insights, it is important to note that our findings are based on a relatively small dataset of 510 headlines. Further research with a larger and more diverse dataset is needed to confirm these findings and to obtain a better understanding of why certain observations occur. Furthermore, since we constrained the length to a limited length of non-stopwords obtained from our study of over 140,000 news headlines length, we are unable to inform about whether the morphology e.g. the length of unconstrained LLM-generated news headlines matches real headlines. A stylometry analysis, whether manual or automated, might provide valuable insights into the patterns and characteristics of the generated news headlines, shedding light on their potential homogeneity and artificial nature. This examination would help determine whether the headlines exhibit robotic tendencies, indicative of being generated in a consistent yet synthetic style. We are currently unable to undertake this examination.

**Declaration of generative AI and AI-assisted technologies in the writing process**

During the preparation of this work the author(s) used Claude.ai, ChatGPT to refine constructs where the content seemed a bit clumsy. In After using this tool/service, the authors reviewed and edited the content as needed and take full responsibility for the content of the publication.

**Funding:** This research received no external funding.

**Data Availability Statement**: The LLM-generated data used in this research is not made available.

**Acknowledgments**:

**Conflicts of Interest**: The authors declare no conflict of interest.

Appendix A1

Table 1: Number of generated categories and headlines

| Category | Generated (m) | Unique (n) |
|---|---|---|
| Academia | 5 | 5 |
| Africa | 5 | 5 |
| Americas | 5 | 5 |
| Arts & Culture | 5 | 5 |
| Asia | 5 | 5 |
| Auto | 10 | 10 |
| Business | 15 | 14 |
| Celebrity | 10 | 8 |
| Comics | 10 | 10 |
| Cooking | 10 | 10 |
| Crime | 10 | 10 |
| Disability Rights | 5 | 5 |
| DIY | 10 | 10 |
| Editorials | 10 | 10 |
| Education | 15 | 15 |
| Entertainment | 10 | 10 |
| Environment | 5 | 5 |
| Europe | 5 | 5 |
| Feature | 10 | 10 |
| Food Security | 5 | 5 |
| Frontpage | 10 | 8 |
| Gardening | 10 | 10 |
| Global Economy | 5 | 5 |
| Global Environment | 5 | 4 |
| Global Health | 5 | 5 |
| Global Politics | 5 | 4 |
| Government | 10 | 3 |
| Headline | 10 | 7 |
| Health | 15 | 15 |
| Home | 10 | 10 |
| Journalism | 10 | 10 |
| Lifestyle | 10 | 10 |
| Local | 10 | 6 |
| Middle East | 5 | 5 |
| National | 10 | 10 |
| News | 10 | 3 |
| Obituaries | 10 | 10 |
| Oceania | 5 | 5 |
| Organizations | 10 | 4 |
| Philanthropy | 5 | 5 |
| Politics | 15 | 13 |
| Press Releases | 10 | 10 |
| Public Events | 10 | 10 |
| Real Estate | 5 | 5 |
| Regional | 10 | 8 |
| Rural Development | 5 | 5 |
| Science | 10 | 10 |
| Social | 10 | 9 |
| Social Justice | 10 | 10 |
| Society | 10 | 10 |
| Space Exploration | 5 | 5 |
| Sports | 15 | 15 |
| Technology | 15 | 12 |
| Tourism | 10 | 9 |
| Transportation | 5 | 5 |
| Travel | 10 | 10 |
| Weather | 10 | 10 |
| World | 10 | 10 |
| Total | 510 | 467 |

Appendix A2

Table: Categories and headline count for the human-written news dataset

| **Category** | **Headline** |
|---|---|
| Accident And Emergency Incident | 92 |
| Accomplishment | 83 |
| Act Of Terror | 99 |
| Animal | 99 |
| Anniversary | 96 |
| Armed Conflict | 97 |
| Arts And Entertainment | 92 |
| Biomedical Science | 85 |
| Bodybuilding | 94 |
| Business Information | 93 |
| Ceremony | 96 |
| Civil Unrest | 98 |
| Climate Change | 98 |
| Communities | 99 |
| Competition Discipline | 96 |
| Conservation | 88 |
| Coup D'etat | 98 |
| Crime | 100 |
| Culture | 86 |
| Demographics | 96 |
| Disaster | 96 |
| Disciplinary Action In Sport | 99 |
| Discrimination | 95 |
| Diseases And Conditions | 91 |
| Drug Use In Sport | 96 |
| Economic Sector | 93 |
| Economy | 94 |
| Election | 99 |
| Emergency Incident | 98 |
| Emergency Planning | 97 |
| Emergency Response | 93 |
| Emigration | 97 |
| Employment | 37 |
| Employment Legislation | 69 |
| Environmental Politics | 98 |
| Environmental Pollution | 95 |
| Exercise And Fitness | 95 |
| Family | 98 |
| Fundamental Rights | 95 |
| Government | 93 |
| Government Policy | 99 |
| Health Facility | 97 |
| Health Organisations | 97 |
| Health Treatment | 96 |
| Healthcare Policy | 99 |
| Immigration | 96 |
| International Relations | 100 |
| Interreligious Dialogue | 88 |
| Judiciary | 91 |
| Justice | 100 |
| Labour Market | 86 |
| Labour Relations | 97 |
| Law | 94 |
| Law Enforcement | 97 |
| Leisure | 97 |
| Lifestyle | 95 |
| Mankind | 98 |
| Market And Exchange | 97 |
| Mass Media | 90 |
| Massacre | 96 |
| Mathematics | 97 |
| Medical Profession | 99 |
| Natural Resources | 97 |
| Natural Science | 94 |
| Nature | 98 |
| Non-Governmental Organisation | 96 |
| Non-Human Diseases | 100 |
| Parent Organisation | 88 |
| Peace Process | 100 |
| People | 100 |
| Plant | 99 |
| Political Crisis | 97 |
| Political Dissent | 97 |
| Political Process | 98 |
| Post-War Reconstruction | 88 |
| Prisoners Of War | 100 |
| Religious Belief | 90 |
| Religious Conflict | 64 |
| Religious Education | 87 |
| Religious Event | 95 |
| Religious Facilities | 95 |
| Religious Institutions And State Relations | 88 |
| Religious Leader | 98 |
| Religious Text | 94 |
| Retirement | 100 |
| School | 97 |
| Scientific Institution | 94 |
| Scientific Research | 99 |
| Scientific Standards | 97 |
| Social Condition | 91 |
| Social Learning | 89 |
| Social Problem | 99 |
| Social Sciences | 92 |
| Sport Event | 98 |
| Sport Industry | 99 |
| Sport Organisation | 100 |
| Sport Venue | 99 |
| Teaching And Learning | 74 |
| Technology And Engineering | 97 |
| Transfer | 82 |
| Unemployment | 98 |
| Unions | 98 |
| Values | 100 |
| Vocational Education | 105 |
| Weather Forecast | 78 |
| Weather Phenomena | 99 |
| Weather Statistic | 94 |
| Weather Warning | 99 |
| Welfare | 100 |
| **Grand Total** | **10260** |

Appendix A3

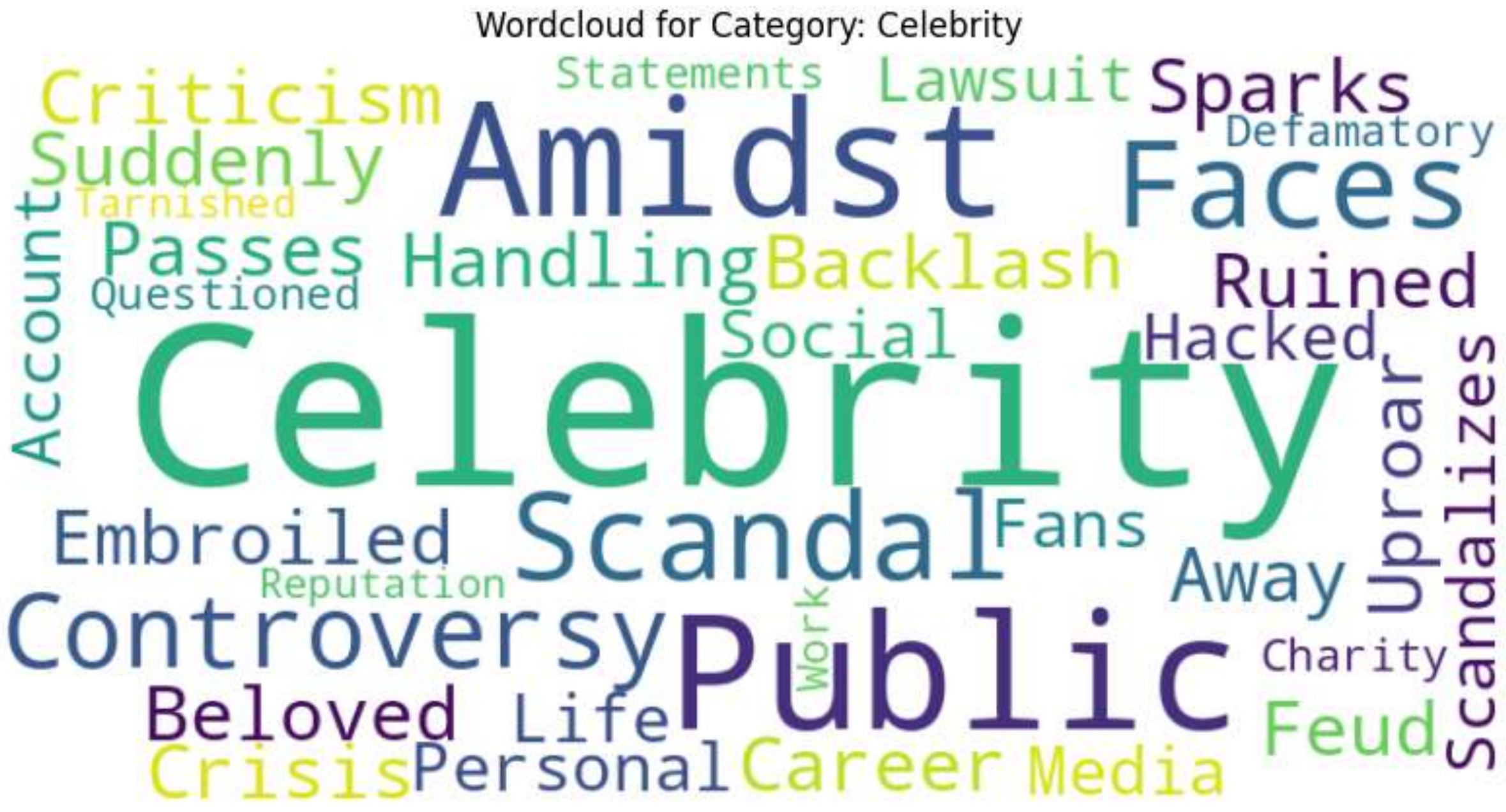

Figure A3.1a. Wordcloud for Celebrity category

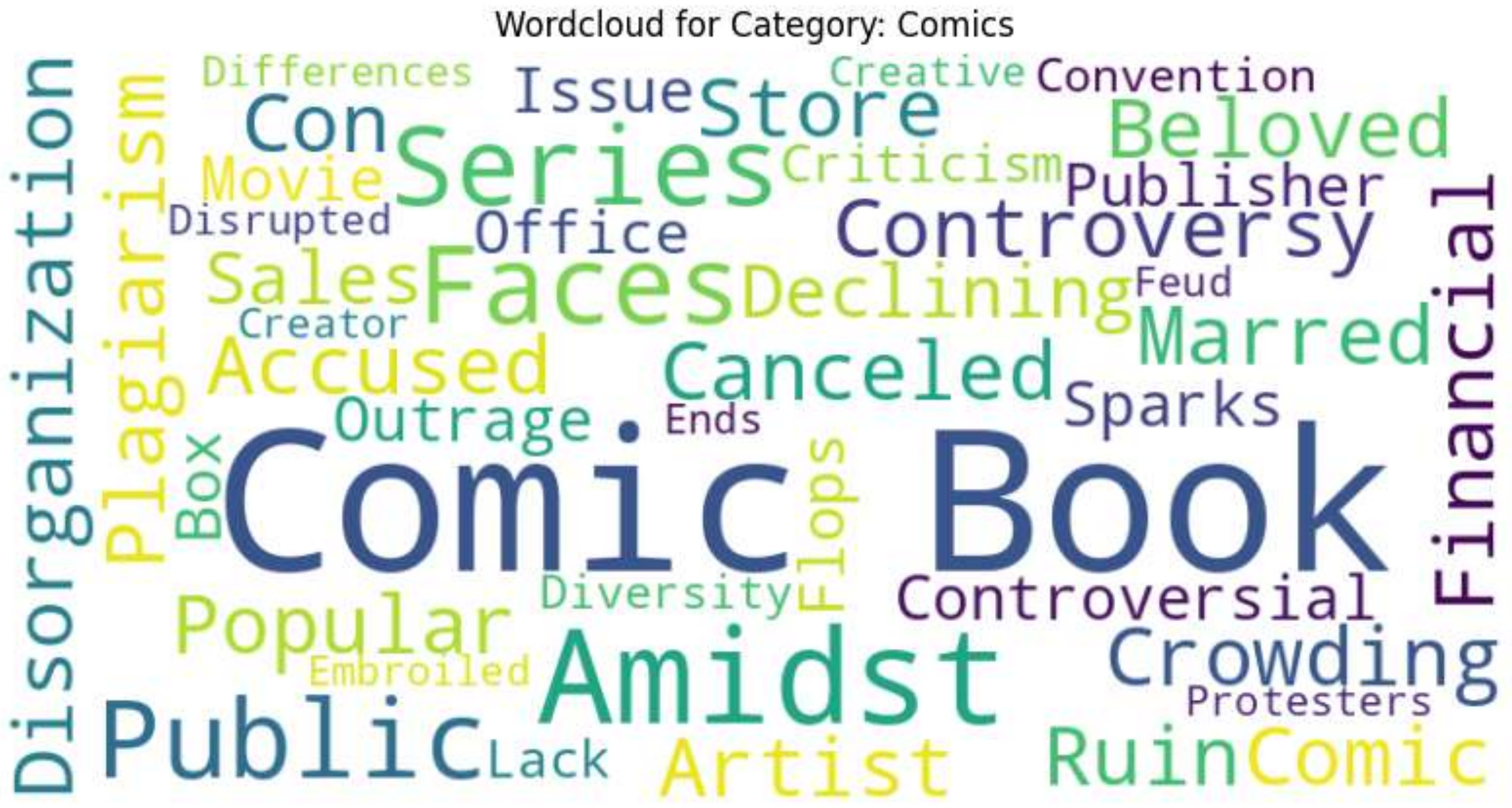

b. Wordcloud for Comics category

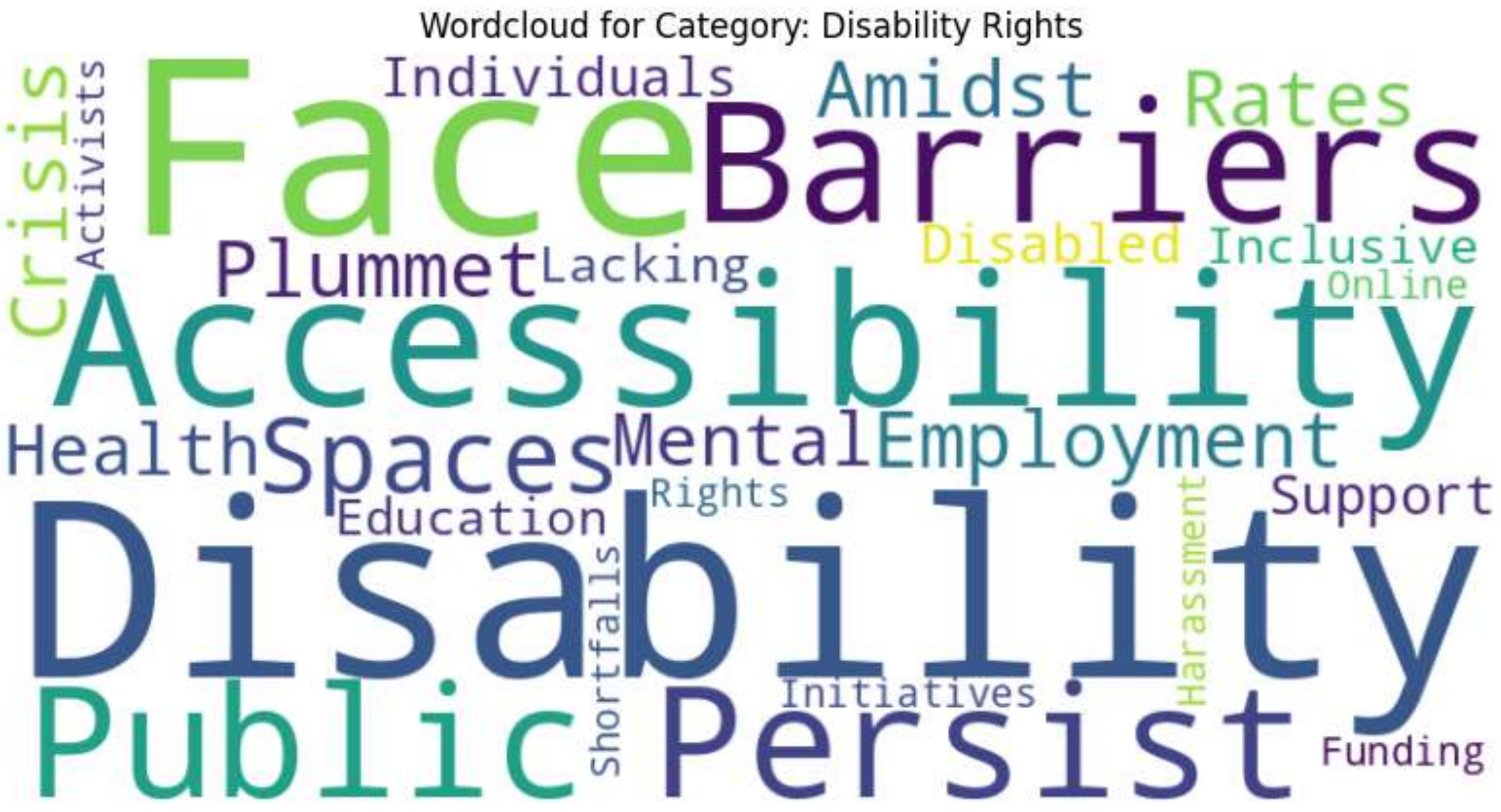

c. Wordcloud for Disability Rights category

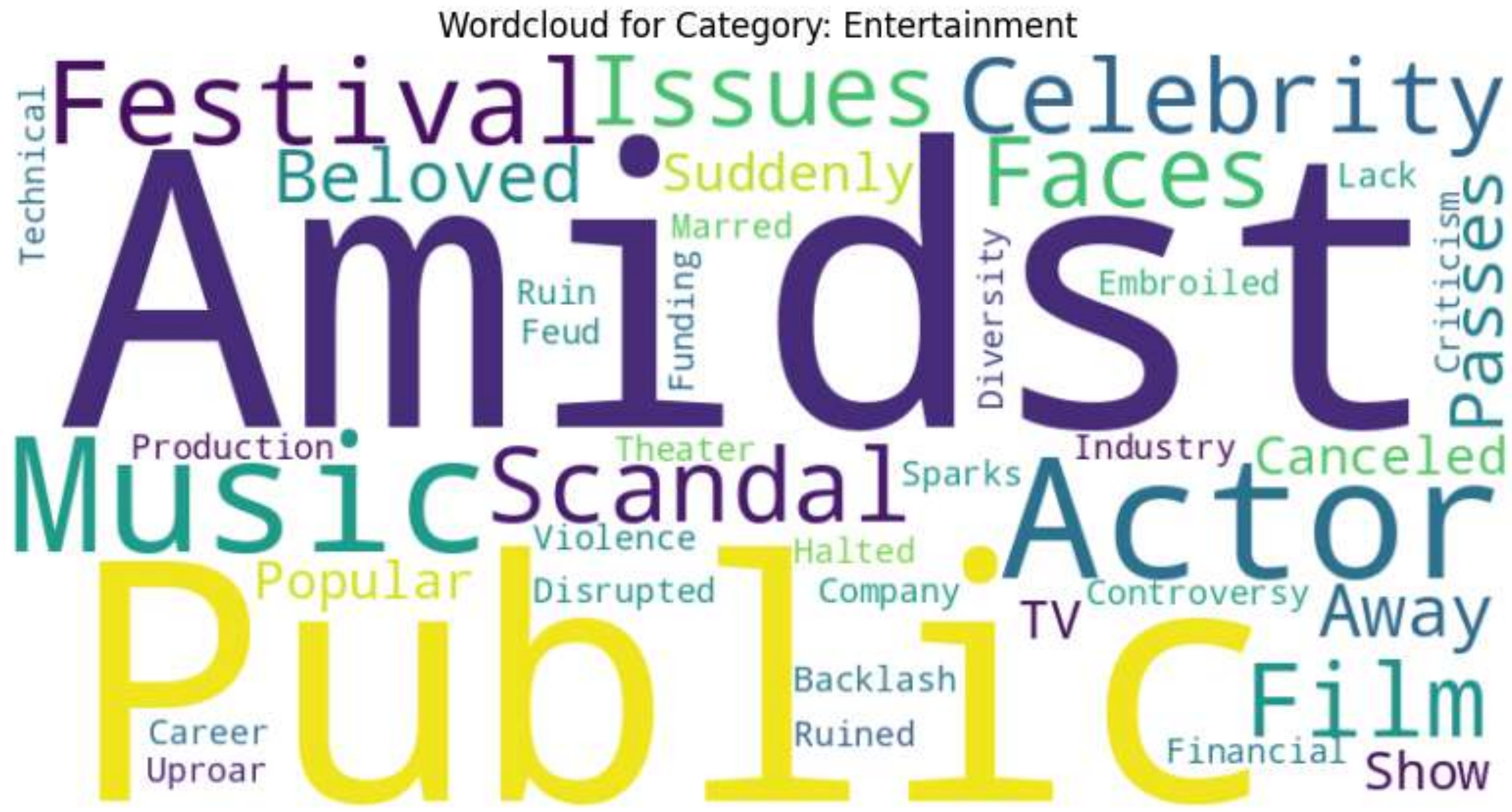

d. Wordcloud for Entertainment category

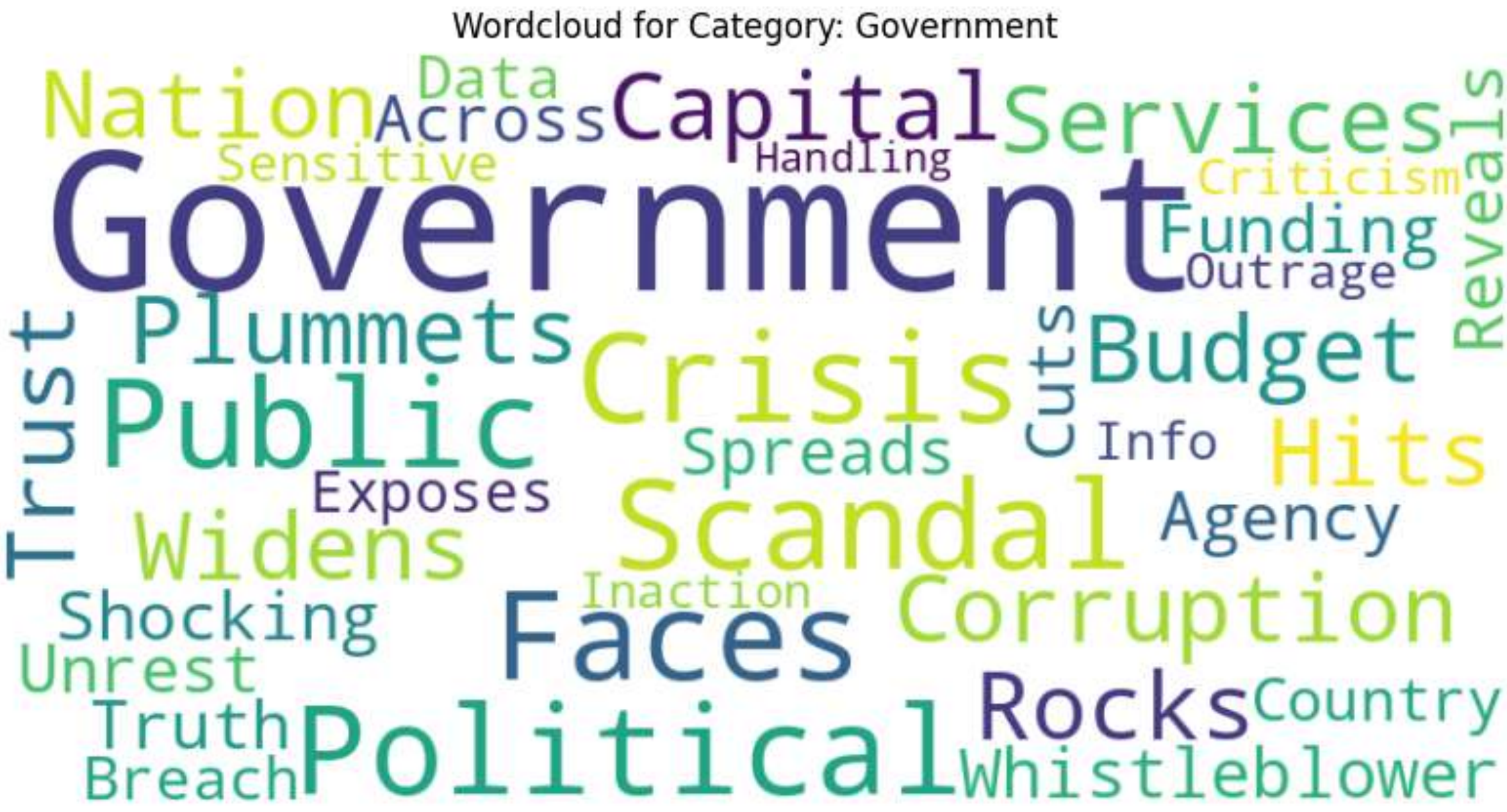

e. Wordcloud for Government category

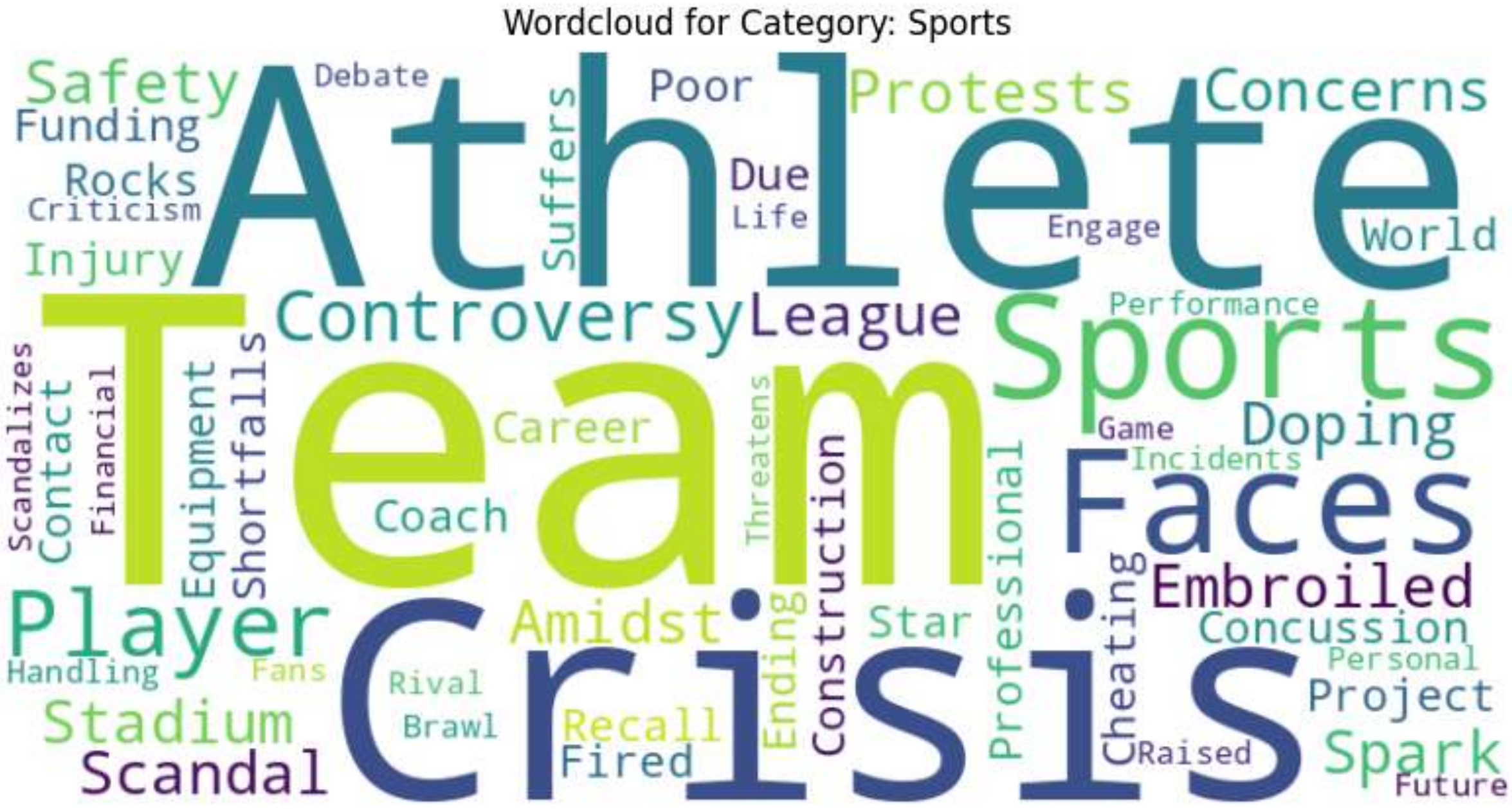

f. Wordcloud for Sports category

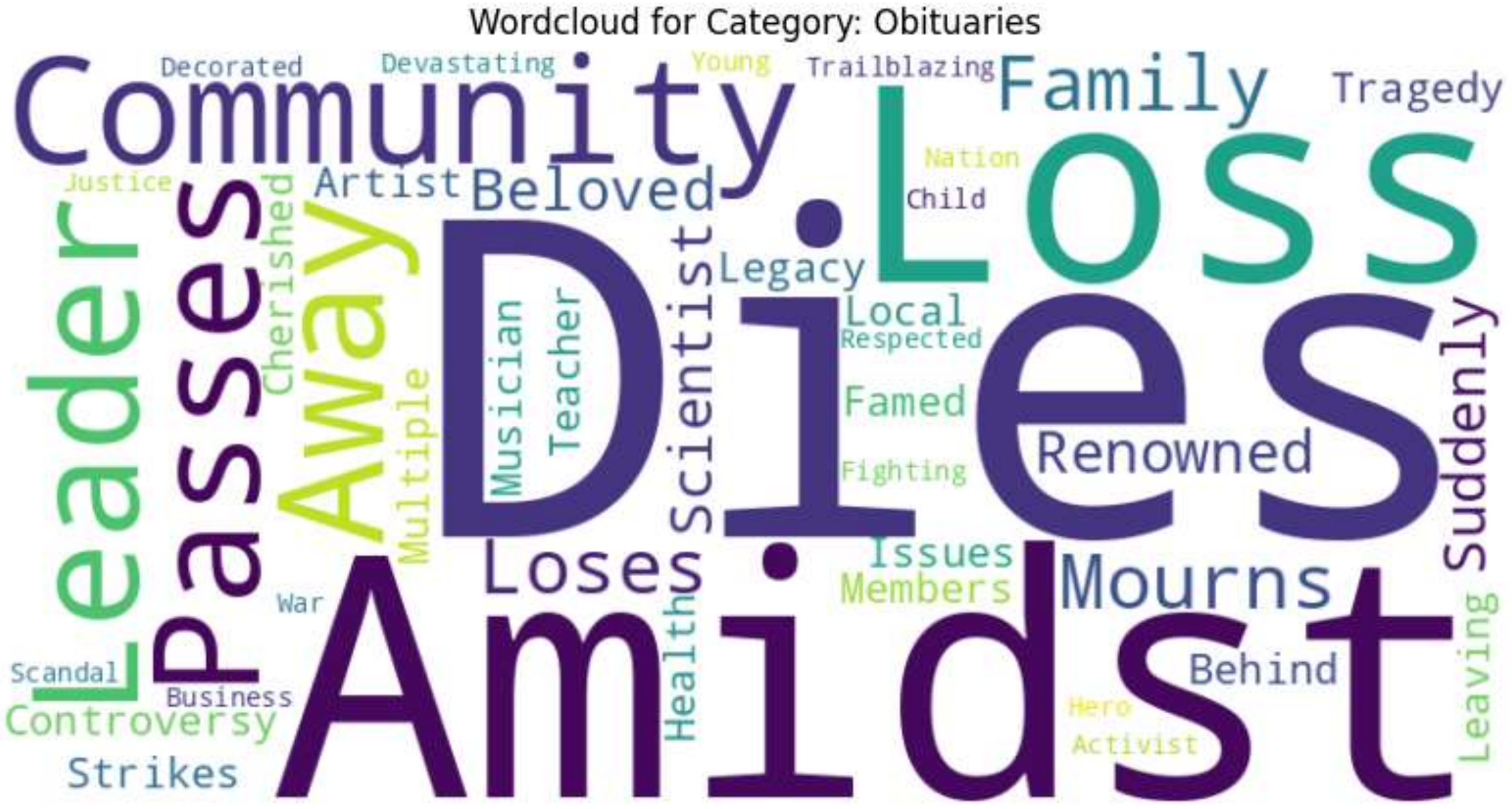

g. Wordcloud for Obituaries category

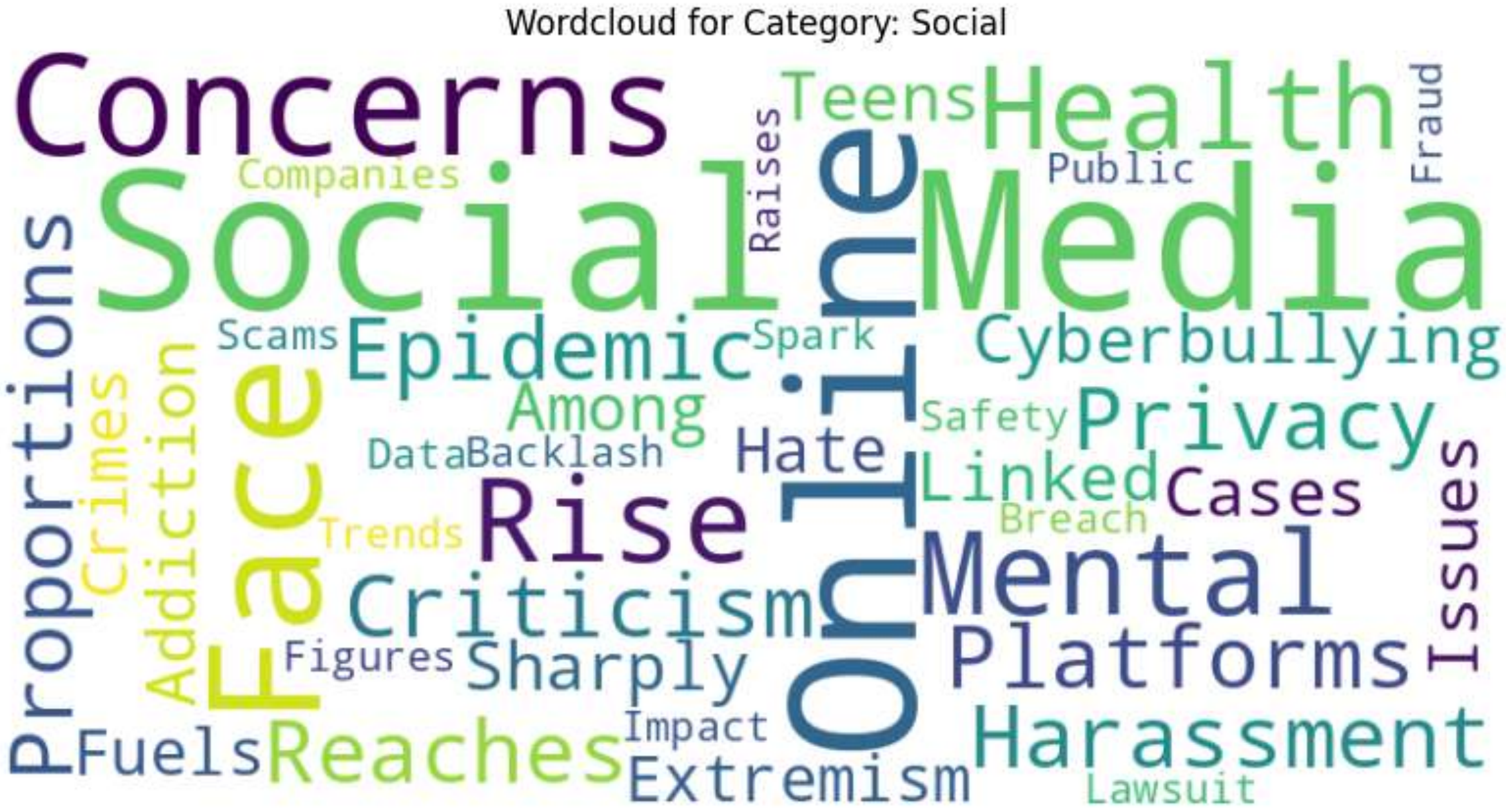

h. Wordcloud for Social category

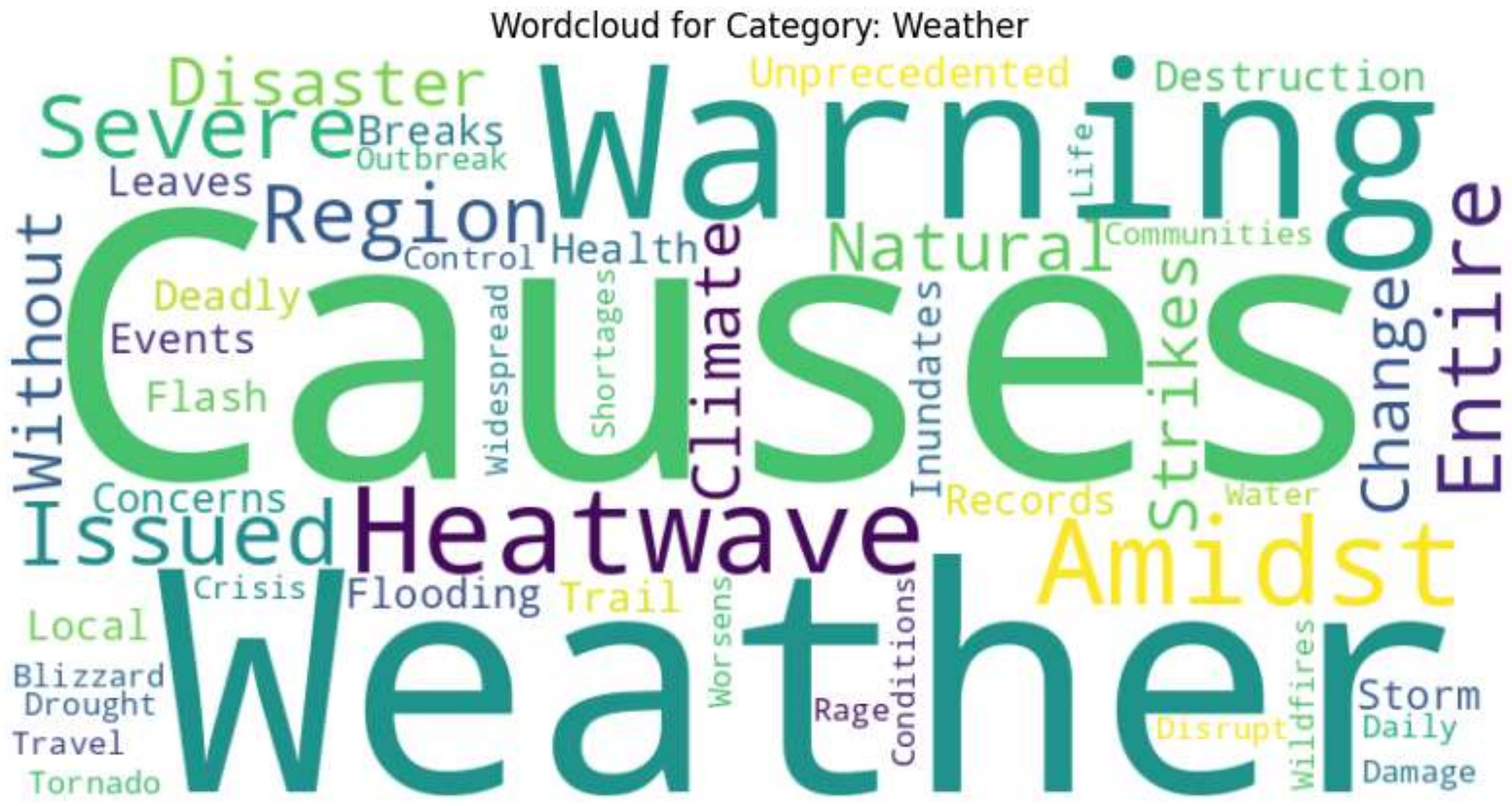

i. Wordcloud for Weather category